\documentclass{article} 
\usepackage{acl}[final]

\usepackage{url}
\usepackage{soul}
\usepackage{multirow}
\usepackage{microtype}
\usepackage{hyperref}
\usepackage{algorithm}
\usepackage{algpseudocode}
\usepackage{booktabs}
\usepackage{amsmath}
\usepackage{graphicx} 
\usepackage[utf8x]{inputenc} 
\usepackage{bbm} 
\usepackage{booktabs}
\usepackage{multirow}
\usepackage{colortbl}
\usepackage[most]{tcolorbox}
\usepackage{amsmath}
\usepackage{enumitem}
\newcommand{\fixchen}[1]{\footnote{\textcolor{red}{\textbf{FIX-CHENG!!!} #1}}}
\newcommand{\fixme}[1]{\footnote{\textcolor{red}{\textbf{FIXME!!!} #1}}}
\newcommand{\OurMODEL}{\textsc{Prompt-SAW}}
\newcommand{\OurDATA}{\textsc{GSM8K-aug}}
\newcommand{\eat}[1]{}
\newcommand{\warn}[1]{\textcolor{red}{#1}}

\newcommand{\li}[1]{\textcolor{red}{#1}}

\DeclareUnicodeCharacter{0301}{\hspace{-1ex}\'{ }}
\newcommand{\di}[1]{\textcolor{green}{#1}}

\title{\OurMODEL{}: Leveraging Relation-Aware Graphs for Textual Prompt Compression}

\author{Muhammad Asif Ali\thanks{The first two authors contributed equally to this work.}$^{*,1,2}$, Zhengping Li$^{*,1,2,4}$, Shu Yang$^{1,2,6}$, Keyuan Cheng$^{1,2,4}$, Yang Cao$^{1,2,4}$\\
\textbf{Tianhao Huang$^{1,2,7}$, Guimin Hu$^{8}$, Weimin Lyu$^{9}$, Lijie Hu$^{1,2,3}$, Lu Yu$^{5} $, and Di Wang$^{1,2,3}$}\\
$^1$Provable Responsible AI and Data Analytics (PRADA) Lab\\
$^2$King Abdullah University of Science and Technology \quad $^3$SDAIA-KAUST AI\\
$^4$South China University of Technology \quad $^5$Ant Group \quad $^6$University of Macau\\
$^7$Nankai University \quad $^8$University of Copenhagen \quad $^9$Stony Brook University
}

\DeclareUnicodeCharacter{0301}{\hspace{-1ex}\'{ }}

\begin{document}

\maketitle

\begin{abstract}
Large Language Models (LLMs) have shown exceptional abilities 
for multiple different natural language processing tasks. While 
prompting is a crucial tool for LLM inference, we observe that 
there is a significant cost associated with exceedingly lengthy 
prompts. Existing attempts to compress lengthy prompts lead to 
sub-standard results in terms of readability/interpretability
of the compressed prompt, with a detrimental impact on prompt 
utility. To address this, we propose~\OurMODEL{}: \textsc{\underline{\textbf{Prompt}}} compres\underline{\textbf{S}}ion 
via Relation \underline{\textbf{AW}}are graphs, an effective strategy 
for prompt compression over task-agnostic and task-aware prompts. \OurMODEL{} uses the prompt's textual information to build a graph, 
later extracts key information elements in the graph to come up 
with the compressed prompt. We also propose~\OurDATA{}, \emph{i.e.,} an extended version 
of the existing GSM8K benchmark for task-agnostic prompts in order to provide a comprehensive evaluation platform. Experimental evaluation using benchmark datasets shows that prompts compressed by~\OurMODEL{} are not only better in terms of readability, 
but they also outperform the best-performing baseline models by 
up to 10.1\% and 77.1\% respectively for task-agnostic and 
task-aware settings while compressing the original prompt 
text by 34.9\% and 56.7\%.

\end{abstract}

\eat{While prompting is a crucial tool for LLM inference, we 
observe the LLMs inference is computationally-inefficient are limited by 
exceedingly lengthy prompts.}

\vspace{-1.7ex}
\section{Introduction}
\label{sec:intro}
\vspace{-0.7ex}
LLMs have attracted considerable attention for their superior performance across a wide range of applications. For this, instructions (aka. prompts) play a crucial role in extending the 
capabilities of LLMs for multiple different tasks. 
The prompts provide the provision to guide the model to elucidate desired model behavior without perturbing the model parameters. This is also highlighted in recent studies that show well-designed 
prompts and the integration of external knowledge are significant to enhance the effectiveness of LLMs'~\citep{PromptEngineeringSurvey}. Different LLMs-related techniques directly benefiting from prompts include but are not limited to: 
In-Context Learning~\citep{ICL}, Chain-of-Thought~\citep{CoT}, 
Retrieval Augmented Generation~\citep{RAG}, and Agents~\citep{Agent} \emph{etc.} 
\eat{play a crucial role in this process.} {Generally, prompts may be sub-divided into two types: task-aware and 
task-agnostic prompts, a quick overview is given in Appendix~\ref{Appendix:aware-prompts} 
and Appendix~\ref{Appendix:agnositic-prompts} respectively.}

At the same time, the abilities of LLMs are significantly compromised/constrained by increasingly lengthy prompts, even comprising thousands of tokens. Lengthy prompts not only obscure requisite information but also increase computational costs and incur inference latency. To tackle this challenge, \textit{prompt compression} techniques, \emph{e.g.,}~\cite{selectiveContext}, have garnered significant interest. These approaches are based on the fact that natural language is inherently redundant~\citep{rongyu}. Thus, it is possible to substantially compress the length of original textual prompts by preserving requisite information in small segments.

Existing prompt compression approaches focus on compressing text at the token level, \emph{i.e.,} they verify whether compression is applicable to each individual token. For instance,~\cite{selectiveContext} proposed Selective-Context that uses a compact language model to evaluate context's lexical units, enabling compression by eliminating units with minimal information. Also, LLMLingua~\citep{llmlingua} and LongLLMLingua~\citep{longllmlingua} developed budget control mechanisms to compresses prompts based on their perplexity.

\begin{figure}[t]
    \centering
    \includegraphics[width=1.0\linewidth]{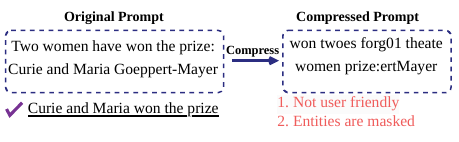}
    \vspace{-5.7ex}
    \caption{An example of compressed prompt compressed using previous token-level based method LongLLMlingua~\citep{longllmlingua}}
    \vspace{-3.7ex}
    \label{fig:example1}
\end{figure}

While existing approaches could enhance the ability to deal with lengthy prompts for LLMs, they lack grammatical coherence, \emph{i.e.,} existing approaches neglect the syntactic and semantic structure of the compressed prompt. 
This is because contemporary prompt compression methods primarily focus on quantifying token-level information, neglecting the overall grammatical structure of the compressed prompt. Such ignorance not only increases the risk of semantic loss within the compressed prompt but also hampers its readability for human readers. An example in this regard is shown in Figure~\ref{fig:example1}, where the original prompt text: {''\em Two women have won the prize: Curie and Maria Goeppert-Mayer''} is compressed to: {\em ''won twoes forg01 theate women prize:ertMayer''}
by LongLLMlingua~\citep{longllmlingua}. 

\eat{By evaluation benchmark, we imply the absence of a universally accepted benchmark for rigorous performance evaluation of the prompt compression techniques. 
The performance of different compression algorithms varies significantly 
in terms of the outcomes of compressed prompts.
Also, the range of original prompts selected by these algorithms is restricted, 
resulting in biased outcomes. 
Furthermore, the absence of standardized benchmark prompts across algorithms 
amplifies the issue of unfair comparisons.}


To fill in the gap, in this paper, we propose~\OurMODEL{},
\emph{i.e.,} 
\textsc{\underline{\textbf{Prompt}}} 
compres\underline{\textbf{S}}ion 
via Relation \underline{\textbf{AW}}are graphs, 
a novel method designed to cut down unnecessary information in 
the prompt text by using Knowledge Graph (KG) structures to exploit 
the small-scale information elements (Section \ref{sec:notation}) 
in the prompts, \emph{i.e.,} information units comprising entities 
and their underlying relations.

\eat{Their succinct and clear representation of information positions Knowledge Graphs (KGs) as an integral component for enhancing LLM, heralding a significant and promising research avenue.}
\OurMODEL{} first extracts all entities and their relations in 
the prompt to formulate the graph.
Later, (i) for task-aware prompts,~\OurMODEL{} looks for
small-scale information elements in the graph to only 
retain task-specific information as a sub-graph, 
(ii) for task-agnostic prompts,~\OurMODEL{} measures
similarity scores between successive information elements 
in the graph to remove the redundant elements to obtain 
required sub-graph.
To retain the syntactic and semantics of the prompt structure, 
\OurMODEL{} finally reinstates the information contained in 
the sub-graph resulting in an optimized and compressed prompt.

We conducted extensive experimental analysis of~\OurMODEL{} 
under both task-agnostic and task-aware settings against 
existing best-performing models as baselines.
For evaluation, we used: (i) ~\OurDATA{}, \emph{i.e.,} an extended 
experimental setting proposed by us for~\textsc{GSM8K}~\citep{GSM8K}, 
(ii) \textsc{NaturalQuestions}~\citep{LostInMiddle},
and (iii) ShareGPT\footnote{\url{https://sharegpt.com/}}. 
Experimental results show that \OurMODEL{} significantly outperforms other baseline models. We summarize the key contributions of this work as follows:
\eat{i.e.,\OurDATA{} offering extended experimental settings on 
previous\textsc{GSM8K}~\citep{GSM8K} and \textsc{NaturalQuestions}~\citep{LostInMiddle}. 
} 
\begin{itemize}
\itemsep0em 
    \item We propose~\OurMODEL{}, a novel framework crafted for 
    compressing prompts by exploiting graph structures to infer 
    key information elements in the prompt that are helpful for 
    compression.
    
    \item As current benchmarks for task-agnostic prompts lack comprehensive evaluation, we propose~\OurDATA{}, an extended version of the existing GSM8K benchmark for an intensive evaluation of \OurMODEL.
    
    \item We demonstrate the effectiveness of~\OurMODEL{} by 
    comprehensive experiments, showing~\OurMODEL{} attained state-of-the-art performance outperforming baseline models by up to 10.1\% and 77.1\% respectively for task-agnostic and task-aware settings while compressing the original prompt text by 34.9\% and 56.7\%.
\end{itemize}


\eat{They only need to be compressed once, and then they can answer all questions. Based on this, 
Later, it uses question-related similarity for compressing task-aware prompts and internal structural similarity for compressing task-agnostic prompts. it organizes the relational triplets as meaningful and 
semantically coherent sentence structures helpful for identifying 
superfluous information.}

\eat{contemporary prompt compression algorithms primarily focus on 
quantifying token-level information content, neglecting the overall structural 
information of the prompt.}

\eat{For example, RAG task prompt is one of task-aware prompt. For each question, we need to retrieve knowledge from the external document and add the retrieved knowledge to the prompt. For this type of prompt, we need to pay attention to the redundancy of relevance to the question.}

\eat{For example, mathematical reasoning based on ICL prompt is one of Task-agnostic prompt.
Compressed reasoning process examples can be applied to all questions. For this type of prompt, we need to pay attention to the redundancy of the internal structure of the prompt.}

\eat{
Apart from this, we observe compressing longer prompts into short 
concise and meaningful 
text is a challenging task. An effective prompt compression strategy 
has to preserve the requisite semantic content while at the same time 
maintaining the end-performance.
It is very hard to enforce stringent constraints on the
compressed prompt, as overly-concise prompts may 
result in losing requisite semantic contents, eventually 
leading to performance degradation.
Moreover, the discrete nature of the text makes it difficult to use 
back-propagation for effective gradient propagation.}

\eat{\OurMODEL{} attained state-of-the-art performance on open-source data sets, \emph{i.e.,}  
\textsc{GSM8K} and \textsc{BBH}. 
Given the fact
During the experimentation phase, it became clear that a single dataset provided a limited range of tasks and lacked a suitable original prompt for fair comparison. 
We also developed
Consequently, we developed a benchmark comprising initial prompts alongside diverse task scenarios.}

\eat{To address these challenges, in this paper we propose \textbf{K}nowledge 
\textbf{G}raph \textbf{B}ased \textbf{P}rompt \textbf{C}ompression (\OurMODEL), 
a novel method designed for prompt compression through the extraction 
of graph structure information.
\eat{, aiming to resolve the aforementioned issues}
Specifically,~\OurMODEL{} first extracts all possible 
\eat{knowledge graph triples or} relational triplets for 
the information contained in the prompt.
\eat{to eliminate redundant semantic information within the prompt.}
Later, it \eat{\textsc{KGBPC} further} organizes the relational
triplets as meaningful and semantically coherent sentence structures 
helpful for identifying\eat{and removing} superfluous information.
\eat{enhances prompt conciseness by relational triples leveraging graph structure insights.}
In the concluding phase, \textsc{KGBPC} reinstates essential knowledge graph triples, resulting in an optimized and compressed prompt.
}

\eat{Despite their advantages, LLMs sometimes face issues like logical errors and factual inaccuracies, which limits their real-world applicability (\citet{Hallucination},~\citep{Hallucination}).}


\eat{Nevertheless, these methods result in the use of increasingly lengthy prompts, 
even comprising thousands of tokens. Such lengthy prompts not only increase 
computational costs but also risk obscuring essential information, thereby 
diminishing the performance of LLMs.}

\eat{
Existing prompt compression approaches may be divided into 
two major categories: soft prompt compression~\citep{softPrompt} 
and discrete prompt compression~\citep{Jung2023DiscretePC}).
Soft prompt compression employs back-propagation to incorporate information from 
labeled training instances. It suffers from lack of cross-model reusability 
and is ineffective for proprietary LLMs (only accessible via APIs). 
On the other hand, discrete prompt compression exploits the discrete nature 
of the text to design strategies that directly changes/edits the prompt tokens.
For instance,~\cite{selectiveContext} proposed Selective-Context that uses 
a compact language model to evaluate context's lexical units, enabling 
compression by eliminating units with minimal information. 
Also, LLMLingua~(\citet{llmlingua},~\citep{llmlingua}) 
and LongLLMLingua~(\citet{longllmlingua},~\citep{longllmlingua}) 
developed budget control mechanisms to compresses prompts based on their perplexity.
}

\eat{Prompt compression technologies are currently divided into two main categories: soft prompt compression~\cite{softPrompt} and discrete prompt compression~\cite{Jung2023DiscretePC}. Soft prompt compression, achieved by training embeddings inclusive of the original context, suffers from a lack of cross-model reusability and is ineffective when Large Language Models (LLMs) are accessible only via APIs. Conversely, a more promising strategy involves directly compressing discrete prompts that consist of specific tokens from the vocabulary.
~\cite{selectiveContext} propose Selective-Context, employs a compact language model to evaluate the self-information of context's lexical units, enabling compression through the elimination of units with minimal informational value. Subsequently, LLMLingua~\cite{llmlingua} and LongLLMLingua~\cite{longllmlingua} developed an innovative budget control mechanism and compresses prompts based on their perplexity.} 

\vspace{-1.7ex}
\section{Related Work}
\label{sec:RL}
\vspace{-0.7ex}
{\bf Prompt Compression.}
Prompt compression techniques are used to reduce the inference 
cost of  LLMs across a wide range of applications.
Existing work can be categorized into 
soft prompt compression and discrete prompt compression.

Soft prompts were introduced by~\citet{softPrompt}. A soft prompt integrates additional trainable parameters at the model's input stage.
\citet{softpromptcompression} emphasized that soft prompt 
compression effectively retains crucial abstract information 
with a reduced parameter count.
\citet{CompressThenPropmt} emphasized that carefully crafted 
prompts are helpful in augmenting the end-performance of 
compressed LLMs, also the compressed LLMs are helpful in the prompt 
learning phase. 

Compared to soft prompt compression, discrete prompt compression 
seeks to optimize the effectiveness of prompts via token-level search 
strategies. 
\citet{Jung2023DiscretePC} employed policy networks to eliminate
unnecessary tokens for prompt compression.
\citet{selectiveContext} utilized self-information metrics to identify 
and remove superfluous information in prompts. 
Capitalizing on these 
advancements, \citet{llmlingua} and \citet{longllmlingua} \eat{have}
formulated algorithms for dynamically adjusting compression rates 
across different prompt sections, giving precedence to tokens with 
higher perplexity\eat{due to their substantial influence}. 

Despite the significant advancements achieved by these studies, 
their primary focus lies on token-level compression, neglecting the 
comprehensive graph structure information inherent in the prompt. 
\eat{To the best of our knowledge, our study represents the first 
exploration of prompt compression using graph structure.}

\eat{{\bf Chain-of-Thought (CoT).}
CoT is a novel prompting technique aimed at enhancing 
the reasoning abilities of LLMs~\citep{CoT}.
It incorporates intermediate problem-solving steps 
to improve LLMs' capacity to tackle 
complex reasoning tasks significantly.
\citet{ToT} exploited the CoT technique and developed Tree-of-Thought (ToT) in order to augment the reasoning 
capabilities of LLMs further. \cite{GoT} proposed Graph-of-Thought 
that employs graphical reasoning capabilities to construct 
prompts.}

\noindent{\bf Knowledge Graphs (KGs) for LLM.}
KGs organize information as structured units, \emph{i.e.,} relational 
triplets (explained in Appendix~\ref{sec:BCKGD_KG}), that 
encapsulate a wide variety of entities/concepts along with 
underlying relations~\citep{KGSurvey}. 
\citet{KGLLMSurvey} illustrated multiple different scenarios for 
integration of KGs with LLM for knowledge and data-driven 
bi-directional reasoning.
\eat{namely: KG-augmented LLMs, LLMs augmented KGs, 
showcasing advancements through KG-enhanced LLM, KG-augmented 
LLM, and synergistic LLM and KG methodologies.}
\citet{ReasoningOnGraph} combined LLMs with KGs for interpretable
reasoning over KG Question Answering tasks.
\citet{KGGPT} introduced an innovative framework that 
leverages LLM's reasoning capabilities for executing KG-based tasks. 
To the best of our knowledge, \OurMODEL{} is the first to
make an attempt to leverage knowledge graph structure for prompt compression.

\begin{figure*}[h]
    \centering
    \includegraphics[width=0.83\linewidth]{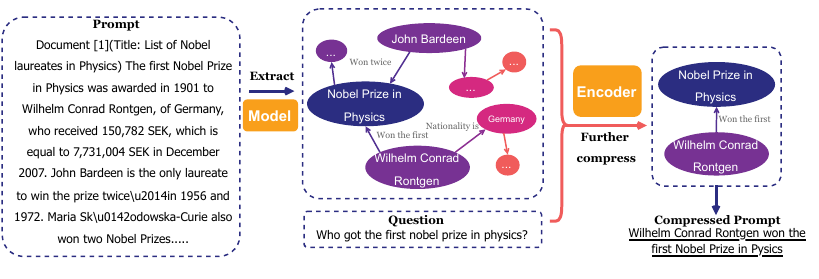}
    \vspace{-2.7ex}
    \caption{Workflow and an example illustration of \OurMODEL{}.}
    \vspace{-3.7ex}
    \label{fig:framework}
\end{figure*}

\eat{
\subsection{Retrieval-Augmented Generation}
Retrieval-Augmented Generation (RAG) integrates pre-trained parametric 
and non-parametric memory to improve the language generation quality of 
LLMs (\citet{RAG},\citep{RAG}).
\warn{By dynamically retrieving pertinent information throughout 
the answer generation phase, the RAG model facilitates the production 
of responses derived from a wider and more accurate knowledge base. 
Recent research identifies RAG as a pivotal technology in enhancing 
the quality and accuracy of LLM.}
\eat{ investigates a model architecture known as, which }
\fixme{We don't need related work on RAG I think. I am omitting it.}}

\eat{
Knowledge Graphs organize structured information as relational triplets 
that encapsulate a vast variety of entities/concepts along with 
underlying relations (\citet{KGSurvey}, \citep{KGSurvey}). 
\cite{KGLLMSurvey} delves into the cutting-edge integration of 
knowledge graphs with LLM, showcasing advancements through KG-enhanced 
LLM, KG-augmented LLM, and synergistic LLM and KG methodologies. 
\cite{ReasoningOnGraph} advocates for the improvement of LLM 
performance in Knowledge Graph Question Answering (KGQA) tasks 
through the innovative generation and retrieval of relational 
paths within Knowledge Graphs. 
\cite{KGGPT} introduces an 
innovative framework that leverages LLM's reasoning capabilities 
for executing Knowledge Graph based tasks. To the best of our 
knowledge, this study represents the inaugural effort to leverage 
knowledge graphs for prompt compression.
}

\eat{To re-emphasized the prompts are widely used to get the best 
possible performance of LLMs across a wide range of application 
scenarios, e.g., Incremental Context Learning~\citep{ICL},
Chain-of-Thought~\citep{CoT}, 
and Retrieval-Augmented Generation~\citep{RAG} etc.,
However, these methods often results in 
wordy/verbose prompts, underscoring the need of 
\eat{research in }prompt compression.}

\vspace{-1.7ex}
\section{Preliminaries}
\label{sec:prelimnaries}
\vspace{-0.7ex}
In this section, we first introduce mathematical notations
and formulate our problem.
Background on the core concepts required for the design 
and development of~\OurMODEL{} is provided in Appendix~\ref{Appendix:Background}.

\subsection{Notations.}
\label{sec:notation}
We use $P$ and $C$ to represent the original and compressed 
prompt respectively.
Likewise, we use $N$ and $\widetilde{N}$ to represent 
the length of the original and compressed prompt. 
We use $\eta=\widetilde{N}/N$ 
to represent the compression rate and $1/\eta$ as the compression ratio.
$\eta^{*}$ is used to represent the target compression rate.
The graph is represented by $\mathcal{G}=\{(e_i,r_i,e_i^{'})\subseteq\mathcal{E}\times\mathcal{R}\times\mathcal{E}\}$,
where $e_i$, $r_i$ and $e_i^{'}$ represent the subject entity, 
relation and object entity in the graph respectively; 
$\mathcal{E} = \{e_1, e_2, \cdots, e_m\}$ and $\mathcal{R}=\{r_1,r_2, \cdots, r_n\}$ denote the set of entities and relations in $\mathcal{G}$.
$g_i = (e_i,r_i,e_i^{'})$ is used to represent small-scale information 
elements in $\mathcal{G}$, equivalent to graph triplet.
$M$ represents auxiliary models used for graph construction.
$E$ is used to represent the encoder network.
$\delta$ represents the similarity threshold used for sub-graph construction.

\subsection{Problem Setting}
In this work, we aim to design and develop an effective prompt compression strategy that can cut down the prompt text by only 
preserving the requisite information content while at the same time 
maintaining the semantics and end performance of the prompt to the best possible extent.

Formally, 
we aim to generate a compressed prompt $C = \{c_i\}_{i=1}^{\widetilde{N}}$
given the original prompt 
$P=(p^\mathrm{ins},p^\mathrm{info},p^\mathrm{que})$,
where $p^\mathrm{ins} = \{p_{i}^\mathrm{ins}\}_{i=1}^{N^\mathrm{ins}}$,
$p^\mathrm{info} = \{p_{i}^\mathrm{info}\}_{i=1}^{N^\mathrm{info}}$, and 
$p^\mathrm{que} = \{p_{i}^\mathrm{que}\}_{i=1}^{N^\mathrm{que}}$, denote
the prompt instruction, information and question, respectively;
$\widetilde{N}$, $N^\mathrm{ins}$, $N^\mathrm{info}$ and $N^\mathrm{que}$ 
represent the number of tokens in $C$, $p^\mathrm{ins}$,
$p^\mathrm{info}$, $p^\mathrm{que}$ and respectively.
We denote $N = N^\mathrm{ins}$ + $N^\mathrm{info}$ + $N^\mathrm{que}$
as the length of the original prompt.


\eat{Following LLMLingua (\citet{llmlingua}, \citep{llmlingua}) and LongLLMlingua (\citet{longllmlingua}, \citep{longllmlingua}), 
we adopt $x=(x^\mathrm{ins},x^\mathrm{info},x^\mathrm{que})$ to represent a prompt $x$, where $x^\mathrm{ins}$ describes the instructions that LLM needs to follow, $x^\mathrm{info}$ contains some additional information such as external knowledge and few-shot demonstrations, $x^\mathrm{que}$ contains the specific questions that the model needs to answer by following the above instructions.}

\eat{Formally, given the original prompt $P = \{p_1,p_2..p_n\}$ with $n$
tokens, we aim to compress it into $m$ tokens, i.e., 
$C = \{c_1,c_2..c_m\}$ with $m\ll n$.
\warn{preserve the semantics of original prompt.}\\
\warn{No loss for the end-application for LLM.}}
\eat{\subsection{Structural Form of Prompt}
We will introduce the specific definition of prompt. According to the different
functions of each part of prompt, it can be divided into different structural parts.}

\vspace{-1.7ex}
\section{\OurMODEL{}}
\label{sec:proposed}
\vspace{-0.7ex}
In this section, we provide details of \OurMODEL{}. 
The workflow is shown in Figure~\ref{fig:framework}.
As shown in the figure, \OurMODEL{} takes the original prompt text as input and generates the compressed 
prompt as the output.

In contrast to the existing token-level compression methods, 
in \OurMODEL{} we use a graph structure to effectively 
represent the textual information in the prompt, which is  helpful to 
analyze the key aspects of the prompt. 
Later, we can refine the information in the graph structure to come up with a compressed prompt in a way that:
(i) The semantic consistency of the compressed prompt is preserved; 
(ii) The end performance and/or utility of the prompt is not distorted.
Below, we first introduce the motivation of \OurMODEL{}, 
followed by the prompt compression process.

\subsection{Motivation of \OurMODEL{}} 
\label{method_structure}
\OurMODEL{} is motivated by the observation that the key information 
within the prompt text could be inferred as a set of entities 
and relations, which can also be organized into a graph structure, 
commonly known as a knowledge graph in literature.

Formally, given a prompt text $P$, we claim it encompasses a 
set of entities
$\mathcal{E}\eat{=\{e_1,e_2,\cdots\}}$, \emph{i.e.,} names of 
persons, locations, 
organizations, miscellaneous elements, etc.,~\citep{ali2020fine}. 
These entities 
serve as the key elements of the prompt structure.
In addition to the entities, we can also infer some relations $\mathcal{R}$ in 
$P$ that may be used to describe the connections between the entities.
\OurMODEL{} re-organizes these key elements of the prompt (\emph{i.e.,} 
entities and their relations) in a graph structure, 
{represented by $\mathcal{G}=
\{(e_i, r_i,e_i^{'})\subseteq\mathcal{E}\times\mathcal{R}\times\mathcal{E}\}$. We use $g_i=(e_i,r_i,e_i^{'})$ to represent the $i$-th information 
element of $\mathcal{G}$, \emph{i.e.,} a fact stating that $e_i$
has $r_i$-$th$ relation with $e_i^{'}$.} 

We argue this transformation of text information to graph is a more reasonable and
natural approach as:
(i) {It helps in highlighting the key information elements in 
the prompt.}
(ii) {Later, analyzing these key entities in combination 
with underlying relations helps in filtering/digging out 
the salient content within the prompt to come up with a
compressed prompt.}

\subsection{Workflow of \OurMODEL{}}
The workflow of \OurMODEL{} consists of two parts. First, it 
uses the information in prompt $P$ to construct a graph $\mathcal{G}$.
Then, based on the specific scenario, we proceed as follows:

\noindent{\bf(a) Task-aware scenario.} For this scenario, 
we traverse the graph $(\mathcal{G})$ in a way to preserve 
only the information elements that are relevant to the task as 
task-specific subgraphs, indicative of information useful 
for the compressed prompt.
\eat{\di{Rewrite this sentence, grammar error}}  \\

\noindent{\bf(b) Task-agnostic scenario.} For this scenario, we no 
longer have access to 
task-specific information. Thus, we use similarity scores 
between the information 
elements in $\mathcal{G}$ to identify and remove the 
redundant elements to obtain 
subgraphs that are helpful for compression. 

Further details about the model components of~\OurMODEL{} 
are provided in the following subsections.  

\subsubsection{Graph Construction.} 
For graph construction from the text data, 
we primarily rely traditional knowledge extraction approaches, \emph{i.e.,} OpenIE~\citep{OpenIE}, to 
construct a graph $G$, as follows

\begin{equation}
   \mathcal{G} = {IE}({P}), 
   \label{eq:Stage:OIE}
\end{equation}

where $P$ is the prompt text, and $IE$ is an 
information extraction module that takes $P$
as input and return graph $G$ as output.
For the cases not addressed by the above equation,
we use an in-context learning 
prompt as our auxiliary method that 
prompts the language model to construct the graph from the original prompt text as follows:
\begin{equation}
   \mathcal{G} = {M}(P_{\text{template}}(\text{P})), 
   \label{eq:graphConstruct}
\end{equation}
where $(P)$ is the prompt text and $P_{\text{template}}$ is the prompt template (explained in Appendix~\ref{Appendix:prompt_Graph}) used 
to guide the LLM $(M)$ to extract the graph $\mathcal{G}$. 
Note, for Equation~\ref{eq:graphConstruct}, we typically prefer a small-scale open-source LLM in order to avoid
higher computational costs incurred by large models.
\eat{, which eventually contributes to the overall computational overhead.}

\subsubsection{Task-aware Prompts}
Task-aware scenarios refer to the settings when the information within the prompt 
is helpful and/or is related to the end-task, \emph{e.g.,} question answering. 
For such cases,~\OurMODEL{} aims to retain only the task-specific information
in $\mathcal{G}$, while filtering out the redundant/useless information.
For this, it first uses an encoder function to get the embeddings for 
the prompt question, as follows.
\begin{equation}
    Emb_{p^{\mathrm{que}}} = E(p^{\mathrm{que}})
\end{equation}
where $Emb_{p^{\mathrm{que}}}$ is the embedding for prompt question 
$(p^{\mathrm{que}})$, and $E$ is the encoder network. 
Then, it computes the pair-wise similarity between the $Emb_{p^{\mathrm{que}}}$ 
and information elements in $\mathcal{G}$, as shown below.
\begin{equation}
    Sim_{\mathcal{G}} = \{E(g_i)\cdot Emb_{p^{\mathrm{que}}} \; | \forall \; g_i \in \mathcal{G} \}
\end{equation}
where $g_i$ corresponds to the $i$-$th$ information element in $\mathcal{G}$, 
$E(g_i)$ is used to encode the information in $g_{i}$, 
$Sim_{\mathcal{G}}$ is the set of the similarity scores between information element 
in $\mathcal{G}$ and the question embeddings $Emb_{p^{\mathrm{que}}}$.
Later, it ranks the scores in $Sim_{\mathcal{G}}$ in order to retain only the elements 
in $\mathcal{G}$ showing a higher degree of similarity with $p^{\mathrm{que}}$, 
as shown below.
\begin{equation}
   Index_{\text{ranked}} = \text{Rank}(Sim_{\mathcal{G}})
\end{equation}
where $\text{Rank}(\cdot)$ is used to sort the similarity scores in 
$Sim_{\mathcal{G}}$ and return corresponding high-ranked information 
elements as $Index_{\text{ranked}}$. 
We then use the information in $Index_{\text{ranked}}$ 
to iterate over $\mathcal{G}$ to extract the sub-graph 
$\mathcal{G}_{\text{subset}}$ 
not surpassing  the targeted compression 
ratio $\eta^{*}$. Its process-flow is illustrated in Algorithm~\ref{alg:alg1}.

\noindent{\bf Workflow of Algorithm~\ref{alg:alg1}.}
The workflow of Algorithm~\ref{alg:alg1} is explained as follows:
(i) initialize $\mathcal{G}_{subset}$ as an empty set (line-1);
(ii) for each element in $Index_{\text{ranked}}$ repeatedly 
add $g_i$ in $\mathcal{G}_{subset}$ until the compression rate
surpasses the target compression rate $\eat^{*}$ (lines 2-7);
(iii) return final graph $\mathcal{G}_{subset}$ as output (line-10).

Finally, we restore/reconstruct the information elements in 
$\mathcal{G}_{\text{subset}}$ to come up with our compressed prompt $C$, as shown below.
\begin{equation}
    C=e_1\oplus r_1 \oplus e_1^\prime ; \cdots ; e_n\oplus r_n \oplus e_n^\prime
   \label{eq:Stage1}
\end{equation}

\eat{
\begin{equation}
    C=e_1\oplus r_1 \oplus e_1^\prime \oplus \text{[SEP]} \oplus \cdots e_i\oplus r_i \oplus e_i^\prime \oplus \text{[SEP]} \oplus\cdots e_n\oplus r_n \oplus e_n^\prime
   \label{eq:Stage1}
\end{equation}
}
where $\oplus$ is the concatenation operator used to combine the entities and 
relations within the information elements $(g_i)$ {in the extracted subgraph $\mathcal{G}_{\text{subset}}$}, and $({;})$ is the delimiter used to separate different information elements in 
$\mathcal{G}_{\text{subset}}$. 

\begin{algorithm}[t]
    \caption{\scshape Subgraph Extraction }
    \label{alg:alg1}
    \begin{algorithmic}[1]
    \Require
        \item[] { \#$\eta^{*}:$ {Target compression rate}}
        \item[] { \#$\mathcal{G}:$ Graph structure of prompt}
        \item[] { \#$len():$ Compute the length of graph structure, as the sum of individual tokens.}
    \Ensure subgraph $\mathcal{G}_{subset}$
        \State $\mathcal{G}_{subset}=\{\}$
        \For{$i \in Index_{\text{ranked}}$}
        \State $\mathcal{G}_{subset}.insert(g_i)$
        \State { \#compute compression rate}
        \State $Rate= len(\mathcal{G}_{subset})/len(\mathcal{G})$
        \State { \#break if meet the constraint}
        \State $\mathbf{If}$ $Rate> \eta^*$ \textbf{then}
        \State $ \qquad Break$
        \State \textbf{end If}
        \EndFor
        \State \Return $\mathcal{G}_{subset}$
    \end{algorithmic}
\end{algorithm}

\subsubsection{Task-agnostic Prompts}

A task-agnostic scenario implies that it is almost impossible to filter useful 
and/or task-specific information within the original prompt text $(P)$.
{In such cases,~\OurMODEL{} looks for recurring information elements in 
$P$ for probable prompt compression.}
We assume two main sources of recurring elements in $P$, \emph{i.e.,} (i) the 
verbose expression of the prompt itself and (ii) the repeated element 
generated by auxiliary models. 
Note that these assumptions are based on empirical observation illustrating 
that large models' re-reading phenomenon leads to the repeated generation of 
the extracted knowledge \citep{Yan2023UnderstandingIL}. 
\eat{\warn{Some examples in this regard are show in Appendix~\ref{Appendix:repeated_know}}.
\fixchen{Add some examples in the Appendix~\ref{Appendix:repeated_know}.}}

For compression over task-agnostic scenarios, we sequentially traverse 
the information elements in $\mathcal{G}$ and select only the elements
exhibiting a lower similarity with priorly selected information 
elements.
Our underlying intuition is that highly similar information elements 
will carry repeated information. Thus, we could avoid redundant information in $P$ by selecting only dissimilar elements.

For this, we use a threshold $\delta$ as a selection criteria for~\OurMODEL{}.
The value of the $\delta$ is determined using a binary search algorithm 
(shown in Algorithm~\ref{alg:alg2}) that computes an appropriate value 
of threshold $\delta$ required to meet the targeted compression rate $\eta^{*}$.

\noindent{\bf Workflow of Algorithm~\ref{alg:alg2}.} The process-flow of Algorithm~\ref{alg:alg2} is explained as follows:
(i) firstly, we initialize an interval $[l,r]$ for the 
threshold $\delta$ (line-1);
(ii) at each step, we partition the interval into two parts $[l, mid]$ and 
$[mid, r]$ via the midpoint $ mid= (l+r)/2$ 
(line-3);
(iii) we will compute the graph subset, \emph{i.e.,} $\mathcal{G}_{subset}$ via function $Compress()$ (explained below) with the value of $mid$ as threshold, shown in line-4;
(iv) compute the compression rate for the $\mathcal{G}_{subset}$ and 
accordingly update the values of $l$ and $r$ (lines 6-9). Specifically, if the compression rate is smaller than $\eta^{*}$, then the current threshold is too stringent thus we judge $\delta$ is in $[mid,r]$, otherwise it is in $[l,mid]$;
(v) depending upon the interval threshold $\gamma$ (line-2),
we compute the value $(l+r)/2$ as the final similarity threshold $\delta$ 
(line-11); 
(vi) finally, use the value of $\delta$ to return the final 
graph subset $\mathcal{G}_{subset}$ (line-13).

\noindent{\bf Compress Function.} The workflow of 
the $Compress()$ is shown in Algorithm \ref{alg:compress} 
and explained as follows: 
(i) start with an empty graph $(\mathcal{G}^\prime)$ (line-1);
(ii) iterate the information elements in the graph $(g_i \in \mathcal{G})$
to compute the similarity score of $g_i$ with the 
elements in $\mathcal{G}^\prime$ to look for maximal similarity, 
\emph{i.e.,} $sim_{max}$ (lines 2-3);
(iii) compare $sim_{max}$ against the compression threshold 
$\delta$ to insert $g_i$ in $(\mathcal{G}^\prime)$ (lines 4-5);
(iv) finally, return $(\mathcal{G}^\prime)$ as the final subset 
of the graph.
The end-goal of Algorithm~\ref{alg:alg2} is to select highly
dis-similar information elements by neglecting cases with
$sim_{max} > \delta$.
For such cases, we assume that the corresponding information element, 
\emph{i.e.,} $g_i = (e_i,r_i,e_i^{'})$ is redundant 
because there is already an element in  $\mathcal{G}^\prime$ that 
is very similar to $g_i$.

\begin{algorithm}[t]
    \caption{\scshape Binary Search}
    \label{alg:alg2}
    \begin{algorithmic}[1]
    \Require
        \item[] { \#$\eta^{*}:$ {Target compression rate}}
        \item[] { \#$\mathcal{G}:$ Graph structure of prompt}
        \item[] { \#$\gamma:$ interval threshold}
    \Ensure subgraph $\mathcal{G}_{subset}$
        \State double $l=0,r=1$
        \While{$r-l > \gamma$}
        \State double $mid=(l+r)/2$
        \State $\mathcal{G}_{subset} = Compress(\mathcal{G},mid)$
        \State $Rate= len(\mathcal{G}_{subset})/len(\mathcal{G})$
        \State $\mathbf{If}$ $Rate > \eta^{*}$ \textbf{then}
        \State $ \qquad r=mid$
        \State \textbf{Else} 
        \State $ \qquad l=mid$
        \EndWhile
        \State $\delta=(l+r)/2$  \# compression threshold
        \State $\mathcal{G}_{subset} = Compress(\mathcal{G},\delta)$
        \State \Return $\mathcal{G}_{subset}$
    \end{algorithmic}
\end{algorithm}

\eat{and populate it over time with selected information elements following 
a similarity threshold.
In order to check the upcoming information elements $(g_i)$ for 
redundancy, we first compute the similarity score of $g_i$ with the 
elements in $\mathcal{G}^\prime$ to look for maximal similarity.
\begin{equation}
    max_{\text{sim}} = \max_{g\in \mathcal{G}^\prime}(E(g)\cdot E(g_i))
   \label{eq:Stage2}
\end{equation}
We use this function to iterate the information elements 
($g_i \in \mathcal{G}$) in order to filter out redundant information.
where $\mathcal{G}^\prime$ is the set of selected }

\vspace{-1.7ex}
\section{GSM8K-AUG}
\label{sec:data_aug}
\vspace{-0.7ex}
As a benchmark dataset, \textsc{GSM8K}~\citep{GSM8K} encompasses high-quality, 
linguistically diverse grade math word problems. However, 
we find that the original dataset poses some limitations 
for evaluating the prompt compression methods. Specifically, it only allows compressing prompts under 
one fix setting, \emph{i.e.,} 8-shot. This is inadequate 
for rigorously evaluating the abilities of the prompt 
compression systems. For instance, it makes it harder 
to analyze and answer the questions: 
(i) Whether prompt compression methods destroy 
connections between individual shots? 
(ii) Also, what impact will these connections have on 
the end-performance for in-context-learning tasks?

To address these limitations, we propose~\OurDATA{},
an extended and more comprehensive experimental setting 
for original GSM8K data. \OurDATA{} extends the original data set 
to $i$-shot setting ($i\in \{1,2,4,8\}$), with 
$i$-shot meaning $i$ example demonstrations in the prompt. 
Note,~\OurDATA{} has a broader coverage, as it encompasses 
the experimental settings of the current GSM8K data settings 
(\emph{i.e.,} 8-shot).
We argue~\OurDATA{} helps in overcoming the limitations mentioned above,
as it provides us with the provision to find correlations between 
different shots.
For instance, it can help us to quickly answer the above 
questions by analyzing the models' performance by compressing 
two prompts at the same time, \emph{i.e.,} 2-shot settings
compared against compressing 
them independently, \emph{i.e.,} 1-shot settings.

\eat{\li{For example, some algorithms may perform well when compressing 1-shot prompt, but perform poorly when the prompt is 8-shot. This means although the algorithm can extract the information of a single shot, it will destroy the overall structure of the prompt, impairs the ICL ability of the target LLM. Comparing prompts of different shots enables us to explore  the impact of prompt compression methods on connections between individual shots further, helping us comprehensively learn its performance on ICL tasks. In addition, for the different shot settings, we can explore the ability of prompt compression methods to find correlations between different shots.}
\di{Why you can answer the above two question? Give more details.}}

\begin{algorithm}[t]
    \caption{\scshape Compress Prompt}
    \label{alg:compress}
    \begin{algorithmic}[1]
    \Require
        \item[] { \#$\delta:$ {Compression threshold}}
        \item[] { \#$\mathcal{G}:$ Graph structure of prompt}
        \item[] { \#$E:$ encoder}
        \item[] { \#$sim():$ function used to calculate similarity}
    \Ensure subgraph $\mathcal{G}^{'}$
        \State $\mathcal{G}^\prime=\{\}$
        \For{$g_i \in \mathcal{G}$}
            \State $sim_{max}=max\{\eat{sim_{max},}sim(E(g),E(g_i)) \;\; \forall g \in \mathcal{G}^\prime\}$
            \If{$sim_{max}<=\delta$}
            \State $\mathcal{G^\prime}.insert(g_i)$
            \EndIf
        \EndFor
        \State \Return $\mathcal{G}^{'}$
    \end{algorithmic}
\end{algorithm}

\vspace{-1.7ex}
\section{Experiments}
\label{sec:exp}
\vspace{-0.7ex}
In this section, we conduct comprehensive experiments for the performance 
evaluation for~\OurMODEL{} compared against different baseline models.

\subsection{Experiment Settings}

\noindent{\bf Datasets.}
To comprehensively evaluate the effectiveness of 
compressed prompts, we evaluate their performance under
both task-agnostic and task-aware data settings. 
For task-agnostic data sets, we consider~\OurDATA{}, 
\emph{i.e.,} an extended variant of the original \textsc{GSM8K}~\citep{GSM8K} 
devised by us to report the model performance  under 
$i$-shot settings with $i\in \{1,2,4,8\}$ (details in Section~\ref{sec:data_aug}). 
For the task-aware dataset, we use NaturalQuestions~\citep{LostInMiddle}, and ShareGPT\footnote{\url{https://sharegpt.com/}}.  
The statistics of dataset is given in 
Table~\ref{tab:DataStat}, and further details 
are provided in Appendix~\ref{Appendix:data}.

\noindent{\bf Baselines.} We compare the performance 
of~\OurMODEL{} against following models as baselines: 
(i) Selective-Context~\citep{selectiveContext},
(ii) LLMLingua~\citep{llmlingua}, 
(iii) LongLLMlingua~\citep{longllmlingua}, and
(iv) GPT4~\citep{GPT4}.
Details about the baselines are provided 
in Appendix~\ref{Appendix:Baseline}.
Note, in order to setup a fair platform for 
comparative evaluation, we re-compute the 
results for the baseline models as per our data 
settings.

\noindent{\bf Evaluation Metrics.} For \OurDATA{}, 
we use Exact Match (EM) as the evaluation
metric. This metric is also employed by 
\citet{GSM8K} and \citet{llmlingua}.
For the evaluation of NaturalQuestions, 
we used Span Accuracy (Span-Acc) as a metric.
This is similar to previous work by~\citet{LostInMiddle} 
and \citet{longllmlingua}. For the evaluation of 
ShareGPT, we used Rouge as the evaluation metric~\citep{rouge_2004}. Apart from these, we also 
use fluency (FL)~\citep{Fluency} to measure the
readability and grammatical coherence of the 
compressed prompt. Further details and mathematical
formulation of these metrics are given in Appendix~\ref{Appendix:Eval}.

\begin{table*}[t]
\centering
\renewcommand{\arraystretch}{1.05} 
\resizebox{0.80\linewidth}{!}{
\begin{tabular}{ccccccccccccccc}
\toprule
\multicolumn{2}{c}{\multirow{2}{*}{Method}} & \multicolumn{3}{c}{\OurDATA{}(1-shot)} & \multicolumn{3}{c}{\OurDATA{}(2-shot)} & \multicolumn{3}{c}{\OurDATA{}(4-shot)} & \multicolumn{3}{c}{\OurDATA{}(8shot)}\\ \cline{3-14} 

\multicolumn{2}{c}{}                        & EM         & Tokens      & 1/$\eta$       & EM         & Tokens      & 1/$\eta$       & EM         & Tokens      & 1/$\eta$      & EM         & Tokens      & 1/$\eta$ \\ \hline
\multicolumn{2}{c}{Original}                & 73.33      & 306         & 1.00      & 78.17      & 612         & 1.00      & 78.92      & 1224        & 1.00     & 82.53      & 2365        & 1.00 \\

\multicolumn{2}{c}{Selective-Context}       & 55.55      & 228         & 1.34      & 58.01      & 449         & 1.36      & 58.13      & 881         & 1.39     & 61.47      & 1752        & 1.35 \\

\multicolumn{2}{c}{LLMLLingua}              & \underline{63.25(7.3\%)}      & 232         & 1.32      & 65.70      & 428         & 1.43      & \underline{67.41(5.5\%)}      & 906         & 1.35     & \underline{73.02(4.2\%)}      & 1799        & 1.31 \\

\multicolumn{2}{c}{GPT4-Generation}         & 60.51      & 215         & 1.42      & \underline{66.44(10.1\%)}      & 411         & 1.49      & 66.39      & 956         & 1.28     & 71.41      & 1273        & 1.86 \\

\multicolumn{2}{c}{\OurMODEL{}}                    & \textbf{67.84}      & 207         & 1.48      & \textbf{73.17}      & 399         & 1.53      & \textbf{71.14}      & 820         & 1.49     & \textbf{76.13}      & 1586        & 1.49  \\ 
\bottomrule[1.0pt]
\end{tabular}
}
\caption{Experimental results on~\OurDATA{}. We report the avg., number of \emph{Tokens} in original and compressed prompts along with EM and compression ratio $(1/\eta)$. For these results, we use GPT3.5-turbo as target LLM. $\eta^{*}$ is set equal to 0.7. We bold-face overall best scores, and
underline the state-of-art along with relative \%-age performance improvement.}
\vspace{-3.1ex}
\label{tab:exp1}
\end{table*}

\noindent{\bf Large Models.}
To demonstrate the generalization of our algorithm 
on different LLMs, we use {GPT3.5-turbo} 
and LLaMA2-7B-chat as our target LLMs. 

\noindent{\bf Experimental Setup.} 
Following the setting of LLMLingua~\citep{llmlingua}, 
we employ greedy decoding with the temperature set to 
0. The max number of tokens generated by LLMs are 
limited to 400.
For graph construction, we use Open-IE tooklit
~\citep{OpenIE} as the primary tool and 
Phi-3-mini~\footnote{\url{https://ollama.com/library/phi3}} as our auxiliary solution.
Note, on average 90-\% graphs were constructed using the 
Open-IE toolkit.
We use OpenAI embedding API~\footnote{\url{https://openai.com/}} as the embedding encoder $(E)$.
The value for $\eta^{*}$ is set to \{0.1, 0.3, 0.5\} for both~ShareGPT and NaturalQuestions, while $\eta^{*}$ = 0.7
for~\OurDATA{}.
In Algorithm~\ref{alg:alg2}, we use $\gamma$ = 0.001. 
All the results reported in 
this paper are averaged over five runs.
All experiments were performed using PyTorch 2.1.0 
with Nvidia RTX 4090 24GB GPU.

\subsection{Experimental Results}

{\bf Results for Task-agnostic Settings.} 
For task-agnostic settings, we report the results of~\OurMODEL{} for~\OurDATA{} in Table~\ref{tab:exp1}.
Note, unlike existing research that reports their 
performance for one fixed setting, we report these results 
for $i$-shot settings, where $i$ indicates the number 
of prompts have been employed by \OurMODEL{}, 
\emph{i.e.,} \{1, 2, 4 and 8\}-shots.

Comparing these results against the baseline models, we can observe
that~\OurMODEL{} outperforms the previous state-of-the-art by a 
significant margin. For instance, compared to the best 
performing baselines, \OurMODEL{} improves the {EM} score by 
up to 7.3\%, 10.1\%, 5.5\% and 4.2\% under 1-shot, 2-shot, 
4-shot and 8-shot settings, respectively.
{Correspondingly reduction in the prompt size is 
32.3\%, 34.9\%, 33.0\% and 32.9\%}.
We attribute such drastic performance improvement to the following factors: 
(1)~\OurMODEL{} retains the logical integrity of the prompts by sub-dividing the original prompts into smaller
comprehensive information elements; 
(2)\OurMODEL{} benefits from the workflow that allows
selecting and omitting individual information elements
for prompt compression without destroying the 
overall information structure of the compressed prompt.
These help~\OurMODEL{} to ensure the utility 
of the compressed prompt for the end task.

\begin{table*}[t]
\centering
\resizebox{0.85\linewidth}{!}{
\begin{tabular}{ccccccccccccc}
\hline
\multicolumn{2}{c}{\multirow{3}{*}{Target LLM}}        & \multicolumn{2}{c}{\multirow{3}{*}{Method}} & \multicolumn{9}{c}{NaturalQuestions}                                                                                                                          \\ \cline{5-13} 
\multicolumn{2}{c}{}                                   & \multicolumn{2}{c}{}                        & \multicolumn{3}{c}{$\eta^{*}$ = 0.5}                  & \multicolumn{3}{c}{$\eta^{*}$ = 0.3}                  & \multicolumn{3}{c}{$\eta^{*}$ = 0.1}                    \\ \cline{5-13} 
\multicolumn{2}{c}{}                                   & \multicolumn{2}{c}{}                        & Acc                                & Tokens & 1/$\eta$  & Acc                                & Tokens & 1/$\eta$  & Acc                                & Tokens & 1/$\eta$   \\ 
\hline
\multicolumn{2}{c}{\multirow{4}{*}{GPT3.5-turbo}} 
& \multicolumn{2}{c}{Original}                & 92.18                              & 524    & 1.00 & 92.18                              & 524    & 1.00 & 92.18                              & 524    & 1.00  \\
\multicolumn{2}{c}{}                                   
& \multicolumn{2}{c}{Selective-Context}       & 49.23                              & 283    & 1.85 & 45.71                              & 173    & 3.03 & 31.12                              & 68     & 7.69  \\
\multicolumn{2}{c}{}                                   
& \multicolumn{2}{c}{LongLLMlingua}           & \underline{59.65(39.0\%)}                        & 270    & 1.94 & \underline{52.02(40.8\%)}                        & 161    & 3.25 & \underline{47.14(14.7\%)}                        & 57     & 9.21  \\
\multicolumn{2}{c}{}                                   
& \multicolumn{2}{c}{\OurMODEL{}}                    & \textbf{82.93}                     & 227    & 2.31 & \textbf{73.22}                     & 136    & 3.86 & \textbf{54.07}                     & 33     & 16.08 \\ 
\hline
\multicolumn{2}{c}{\multirow{4}{*}{LLaMA2-7B-chat}}    & 
\multicolumn{2}{c}{Original}                & \multicolumn{1}{c}{71.25}          & 524    & 1.00 & \multicolumn{1}{c}{71.25}          & 524    & 1.00 & \multicolumn{1}{c}{71.25}          & 524    & 1.00  \\
\multicolumn{2}{c}{}                                   & 
\multicolumn{2}{c}{Selective-Context}       & \multicolumn{1}{c}{40.87}          & 283    & 1.85 & \multicolumn{1}{c}{34.26}          & 173    & 3.03 & \multicolumn{1}{c}{21.37}          & 68     & 7.69  \\
\multicolumn{2}{c}{}                                   & 
\multicolumn{2}{c}{LongLLMlingua}           & \underline{43.56(77.1\%)}   & 270    & 1.94 & \underline{38.87(71.2\%)}    & 161    & 3.25 & \underline{26.12(72.7\%)}    & 57     & 9.21  \\
\multicolumn{2}{c}{}                                   & 
\multicolumn{2}{c}{\OurMODEL{}}                    & \multicolumn{1}{c}{\textbf{77.14}} & 227    & 2.31 & \multicolumn{1}{c}{\textbf{66.53}} & 136    & 3.86 & \multicolumn{1}{c}{\textbf{45.11}} & 33     & 16.08 \\ 
\hline
\end{tabular}}
\vspace{-1.7ex}
\caption{Experimental results on NaturalQuestions. We bold-face overall best scores, and underline the existing state-of-art along with relative \%-age performance improvements.}
\label{tab:exp2}
\vspace{-2.7ex}
\end{table*}

\renewcommand{\arraystretch}{1.5}

\begin{table*}[]
\centering
\resizebox{\linewidth}{!}{
\begin{tabular}{clclccccccccccccccc}
\hline
\multicolumn{2}{c}{\multirow{4}{*}{Target LLM}}        & \multicolumn{2}{c}{\multirow{4}{*}{Method}} & \multicolumn{15}{c}{\multirow{2}{*}{ShareGPT}}                                                      \\
\multicolumn{2}{c}{}                                   & \multicolumn{2}{c}{}                        & \multicolumn{15}{c}{}                                                                                                                                                                                  \\ \cline{5-19} 
\multicolumn{2}{c}{}                                   & \multicolumn{2}{c}{}                        & \multicolumn{5}{c}{$\eta^{*}$ = 0.5}                                & \multicolumn{5}{c}{$\eta^{*}$ = 0.3}                                & \multicolumn{5}{c}{$\eta^{*}$ = 0.1}                                 \\ \cline{5-19} 
\multicolumn{2}{c}{}                                   & \multicolumn{2}{c}{}                        & Rouge-1         & Rouge-2         & Rouge-L         & Tokens & 1/$\eta$  & Rouge-1         & Rouge-2         & Rouge-L         & Tokens & 1/$\eta$  & Rouge-1         & Rouge-2         & Rouge-L         & Tokens & 1/$\eta$  \\

\hline
\multicolumn{2}{c}{\multirow{3}{*}{GPT3.5-turbo}} & \multicolumn{2}{c}{Selective-Context}       & 36.41          & 16.48          & 23.17          & 312    & 1.80 & 34.41          & 14.81          & 22.04          & 217    & 2.59 & 33.32          & 11.87          & 19.66          & 95     & 5.92 \\

\multicolumn{2}{c}{} & \multicolumn{2}{c}{LongLLMlingua}           & \underline{38.13(29.3\%)}          & \underline{17.07(52.6\%)}          & \underline{25.22(29.1\%)}          & 305    & 1.84 & \underline{36.13(34.9\%)}          & \underline{15.61(70.9\%)}          & \underline{23.17(38.3\%)}          & 202    & 2.78 & \underline{34.23(38.6\%)}         & \underline{12.64(87.6\%)}          & \underline{21.16(32.0\%)}          & 90     & 6.24 \\

\multicolumn{2}{c}{}                                   & \multicolumn{2}{c}{\OurMODEL{}}                    & \textbf{49.31} & \textbf{26.04} & \textbf{32.57} & 176    & 3.19 & \textbf{48.75} & \textbf{26.68} & \textbf{32.04} & 146    & 3.85 & \textbf{47.44} & \textbf{23.71} & \textbf{27.93} & 61     & 9.21 \\ \hline

\multicolumn{2}{c}{\multirow{3}{*}{LLaMA2-7B-chat}}    
& \multicolumn{2}{c}{Selective-Context}       & 32.24          & 13.42          & 23.74          & 312    & 1.80 & 30.78          & 10.95          & 22.96          & 217    & 2.59 & 30.15          & 10.32          & 19.13          & 95     & 5.92 \\

\multicolumn{2}{c}{}                                   
& \multicolumn{2}{c}{LongLLMlingua}           & \underline{34.51(0.8\%)}          & \textbf{15.18} & \textbf{27.73} & 305    & 1.84 & \underline{32.97(1.1\%)}          & \textbf{13.78} & \textbf{26.01} & 202    & 2.78 & \underline{31.26(2.8\%)}          & \textbf{12.11} & \textbf{23.08} & 90     & 6.24 \\

\multicolumn{2}{c}{}                                   
& \multicolumn{2}{c}{\OurMODEL{}}                    & \textbf{34.81} & \underline{14.03(8.2\%)}          & \underline{24.88(11.4\%)}          & 176    & 3.19 & \textbf{33.34} & \underline{12.19(13.0\%)}          & \underline{24.08(8.0\%)}          & 146    & 3.85 & \textbf{32.15} & \underline{11.34(6.8\%)}          & \underline{20.35(13.4\%)}          & 61     & 9.21 \\ \hline
\end{tabular}}
\vspace{-1.7ex}
\caption{Experimental results on ShareGPT. We bold-face overall best scores, and underline the existing state-of-art along with relative \%-age performance improvements.}
\label{tab:exp_sharegpt}
\vspace{-2.7ex}
\end{table*}

\noindent{\bf Results for Task-aware Settings.} Table~\ref{tab:exp2} 
reports the performance of~\OurMODEL{} under task-aware 
settings on  NaturalQuestions using GPT3.5-turbo 
and LLaMA2-7B-chat as target LLMs. These results show that, 
\OurMODEL{} improves the Span Accuracy by 39.0\%, 40.8\% and 14.7\% for GPT3.5-turbo, and 77.1\%, 71.2\%, 72.7\% for LLaMA2-7B-chat, respectively, for different values of the target compression rates $\eta^{*} = \{0.5, 0.3, 0.1\}$, 
against the best-performing baseline (LongLLMLingua).
Correspondingly, the reduction in the prompt size is 
56.7\%, 74.0\%, and 93.7\%\ respectively.

The results of~\OurMODEL{} on ShareGPT (in Table~\ref{tab:exp_sharegpt}) show for GPT3.5-turbo as 
target LLM, \OurMODEL{} improves the Rouge-1 score 
by up to 29.3\%, 34.9\% and 38.6\%. The performance 
for Rouge-2 and Rouge-L exhibit a similar behavior. 
For these results, we also observe for LLaMA2-7B-chat, 
the improvement in performance is relatively lower 
compared to that of GPT3.5-turbo.
A probable justification in this regard is the fact 
that LLaMA2-7B-chat is more influenced by the change in context for the compressed prompt. Whereas, higher performance on GPT3.5-turbo indicates that~\OurMODEL{} preserves the critical information in the prompt.

Correlating the results for both settings, we observe 
that compared to the task-agnostic scenarios,~\OurMODEL{} yields better performance for task-aware settings, 
especially for NaturalQuestions.
This is due to 
the fact that it is more difficult to dig out the latent 
correlation between information elements within the prompt's 
internal structure rather than explicit task-aware 
correlation extraction.

\begin{figure}[b]
\vspace{-3.7ex}
\centering
\begin{tabular}{lll}
\includegraphics[width=0.22\textwidth]{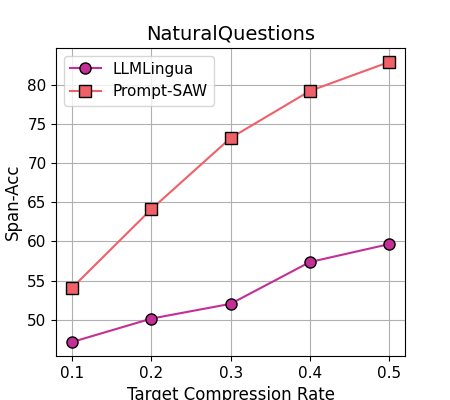} &
\includegraphics[width=0.22\textwidth]{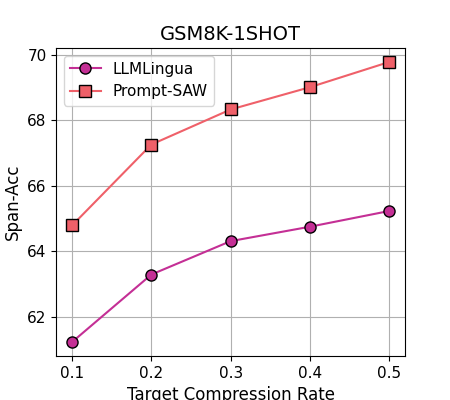} &
\end{tabular}
\vspace{-2.7ex}
\caption{Different target compress rate between LLMLingua and \OurMODEL{} on NaturalQuestions and \OurDATA{} dataset using GPT3.5-turbo as target LLM.}
\label{fig:CompressRate}
\vspace{-3.7ex}
\end{figure}

From Table~\ref{tab:exp2} and Figure~\ref{fig:CompressRate} 
we can also find that the performance of~\OurMODEL{} drops 
when the value for the target compression rate $(\eta^{*})$ 
decreases from 0.5 to 0.1. A probable justification that the 
actual compression ratio of~\OurMODEL{} is significantly 
higher than the target compression ratio when the target 
compression ratio is 10, \emph{i.e.,} $\eta^{*}=0.1$. 
This is owing to the fact that~\OurMODEL{} only retains the information 
elements in $\mathcal{G}$ as the key/basic information units 
for the compression process. It will delete some entities and 
relations that may be highly similar to the problem but their 
overall structure is too long, leading to relatively poor 
performance for a higher compression ratio.

\eat{From Table~\ref{tab:exp_sharegpt} we can find that when using the ShareGPT dataset, utilizing ChatGPT3.5-turbo-0301 as the target LLM can achieve greater performance improvement than using Llama2-7B as the target LLM. A probable justification is that weaker LLMs (\emph{e.g.,} LLaMA2-7B-chat) are more influenced by the context when generating output. The great performance on GPT3.5-turbo indicates that \OurMODEL{} actually does not reduce the critical information from past conversations.}

\eat{which may be 
accredited to a higher compression ratio.}
\eat{Overall, these results show that~\OurMODEL{} exhibits 
better results compared to the baseline models.}

\eat{\textbf{Different target compress rate via LLMs}. As shown in Figure~\ref{fig:CompressRate}, the performance of \OurMODEL{} decreases relatively gently with the decrease of compression rate on \OurDATA{}. [Reason]. The compression rate has a greater impact for \OurMODEL{} on NatrualQuestions. At a compression rate of 0.1, since our compression granularity is not token, but triplet, the actual compression rate is much lower than 0.1. We check the compression results and find that .}

\subsection{Further Analysis}
In this section, we perform an in-depth analysis
of the performance of~\OurMODEL{}. 

\eat{ 
{\bf Computational Overhead.} We partition the overall computational 
overhead of~\OurMODEL{} into compression overhead and subsequent 
inference overhead. This may be computed as follows:
\begin{equation}
    c_{\OurMODEL{}} = \underbrace{(P+N)\cdot c_{\text{small}}+N^{\prime}\cdot
    c_{\text{encoder}}}_{Compressed\ Overhead}+\widetilde{N}\cdot c_{\text{LLMs}},
\end{equation}
where $c_{small}$, $c_{encoders}$, and $c_{LLM}$ represent the per token overhead of the small-scale language \warn{model $M$ in \ref{eq:Stage1}}, encoder $E$, and LLM used to answer, respectively. $P$ 
is the length of prompt text used to extract graph structure, $N$ is the length of the original prompt, $N^{\prime}$ is the total length of elements included in the graph. \di{No analysis and results}
\warn{We also discuss using additional approaches to replace small 
LLMs in Appendix~\ref{Appendix:graph-construct}, which will 
further reduce the compression overhead of our method.}
} 

\eat{\textbf{The Overhead of LLMLingua}\\
\begin{equation}
    c_{LLMLingua}=(L+kL/\tau+L/\tau)\cdot c_{\text{small}}+L/\tau\cdot c_{\text{LLMs}}
\end{equation}
where $c_{small}$ and $c_{LLMs}$ represent the per token overhead of the small LM and LLM. $L$ is the length of original prompt, $kL/\tau$ is the perplexity calculation of tokens, and $L/\tau$ is the conditioned perplexity calculation of compressed results. The token numbers of perplexity and conditioned perplexity are consistent with those mentioned in LLMLingua.}

\eat{We formally discuss the cost of compress algorithm by setting compression threshold $\eta$, which mean the compression ratio $\tau$ is $1/\eta$. \\}

\noindent{\bf Different Target LLMs.}
We also analyze the performance of~\OurMODEL{} using 
different target LLMs. Comparing the results in Table~\ref{tab:exp2} and Table~\ref{tab:exp_sharegpt}, 
we can find that GPT3.5-turbo performs better than 
LLaMA2-7B-chat.

For NaturalQuestions, GPT3.5-turbo achieves up to 19.8\% 
higher Acc scores compared to that of LLaMA2-7B-chat.
Likewise for ShareGPT, it achieve up to 47.55\% higher 
value for Rouge-1. We attribute this result to GPT3.5's 
stronger ability to understand and comprehend context, 
resulting in generating higher quality response for 
the prompts compressed by~\OurMODEL{}.

\eat{However, when we consider the NatrualQuestions dataset, 
using LLaMA-2-7B-chat as target LLM allows \OurMODEL{} to 
achieve greater performance improvements compared to the 
baseline. 

This is because \OurMODEL{} can extract vital information 
from the prompt to help answer questions, which is harder 
for weaker LLM to find.}

\begin{table}[h]
\centering
\scalebox{0.8}{
\begin{tabular}{cccc}
\hline
        & Selective-Context & LLMLingua  & \OurMODEL{} \\ \hline
FL      & 5.61              & 5.74       & \textbf{6.30} \\ \hline  
\end{tabular}}
\vspace{-1.7ex}
\caption{Fluency (FL) of the compressed prompt on \OurDATA{}. We report the performance of~\OurMODEL{} compared against the baseline models. The best scores are bold-faced.}
\label{tab:exp4}
\vspace{-3.7ex}
\end{table}

\noindent{\bf Readability  of Compressed Prompts.}
As explained in the introduction (also highlighted in 
Figure~\ref{fig:example1}), a key limitation of existing  
prompt compression approaches is the limited readability 
of the compressed prompt. In order to validate the results of~\OurMODEL{} in terms
of human readability and/or interpretability, we report some
example prompts along with prompt compressed using~\OurMODEL{}
and LLMLingua~\citep{llmlingua}
in Appendix~\ref{Appendix:exp-readability}, for a quick comparison. These examples clearly indicate that the 
prompt compressed by~\OurMODEL{} exhibit better readability 
and/or interpretability compared to compressed using 
LLMLingua. 

For instance, as shown in Example~\ref{Appendix:ex1} 
(Appendix~\ref{Appendix:exp-readability}),
the prompt compressed by LLMLingua encompasses
grammatical incoherent text, such as:
\{\emph{``List of Nobelates in The first Prize1 
Wilhelmrad, of who received82 in en prize''}\}.
Lack of grammatical coherence significantly undermines 
the readability and/or interpretability of the compressed 
prompt, thus impacting its end-utility.
Whereas, the prompt compressed by~\OurMODEL{}, relying on knowledge graph triples exhibits a consistent grammatical 
sentence structure, as shown in the lower half of Example~\ref{Appendix:ex1}.

To further support our claims, we also conducted a 
quantitative comparison. Specifically, we assess the 
fluency of the compressed prompts through the computation of a weighted mean of bi-gram and tri-gram entropies~\citep{Fluency}. 
Its computational details are given in Appendix~\ref{Appendix:Eval}, and result 
is reported in Table~\ref{tab:exp4}.
These results show that~\OurMODEL{} yields relatively 
higher fluency scores than the baseline models.
A lower score for baseline models, \emph{e.g.,} LLMLingua, is attributable to loss of intrinsic semantic relationship 
between the tokens for the compressed prompt. 



\eat{
\noindent{\bf Fluency of Compressed Prompts.}
\li{We also conducted a quantitative comparison to 
support our claims that the prompts compressed by~\OurMODEL{} are more fluent.} For this, we assess the fluency of the compressed prompts 
through the computation of a weighted mean of bi-gram and tri-gram entropies~\citep{Fluency}. Its computational details are given in Appendix~\ref{Appendix:Eval}, and result 
is reported in Table~\ref{tab:exp4}. \di{This paragraph is strange, it is almots the same the above paragraph. Reorganize it. }
}

\noindent{\bf Computational Overhead.} One of 
the key objectives of prompt compression is 
to efficiently reduce the overall computational and corresponding fiscal cost associated with the proprietary LLMs, while at the same time preserving the end-utility of the prompt.

In order to compute the computational overhead of~\OurMODEL{}, we assume embedding and computing similarity takes a constant amount of time and its cost
may be ignored because it is much smaller than that of LLM.
Specifically, we use the following formulation to study 
the computational efficiency of~\OurMODEL{}:
\vspace{-1.3ex}
\begin{equation}
    c = L \cdot c_{\text{graph}} + (L \cdot \eta^{*}) \cdot c_{\text{LLMs}},
\end{equation}
\vspace{-0.3ex}
where $c_{\text{graph}}$ represents the per-token computation load to generate knowledge triples 
and $c_{\text{LLMs}}$ is per-token computation for target LLMs to generate final output respectively. 
$L$ represents length of the original prompt, $\eta^{*}$ is the target compress rate, and
$c$ represents the total computational overhead for compression.

Following the assumption of \cite{llmlingua}, we estimate $c_{\text{graph}}$ and $c_{\text{LLMs}}$ based on model parameters, as: 
$c_{\text{graph}} \approx 0.3/175c_{\text{LLMs}} \approx 0.0017 \cdot c_{\text{LLMs}}$ using OpenIE toolkit as the knowledge extraction approach. 
When $\eta^{*} = 0.2$, we have $c \approx  0.2017L \cdot c_{\text{LLMs}}$. This means we can achieve nearly 5x savings in terms of computational resources.
For the cases employing a small LLM for graph construction, $c_{\text{graph}} \approx 3.8/175c_{\text{LLMs}} \approx 0.02 \cdot c_{\text{LLMs}}$, the computational overhead is $c \approx 0.22 \cdot c_{\text{LLMs}}$, which is only slightly higher than the first case.

Overall, these results show the computation 
efficiency of our model is comparable with 
that of~\citet{llmlingua}. On the other hand,~\OurMODEL{} 
offers much higher benefits, \emph{i.e.,} compressing
prompts without distorting their readability for 
end-users, while at the same time preserving their 
end-utility to the best possible extent.

\eat{\textbf{task-aware}\\
We then test the performance of~\OurMODEL{} for the task-aware prompt on NaturalQuestions, the results are shown in Table~\ref{tab:exp2}. Compare to Table~\ref{tab:exp1}, the results show that \OurMODEL{} achieves greater improvement than baselines on task-aware prompt. We attribute this to the fact that internal structure correlation extraction is more difficult than problem structure correlation extraction.\\
Although \OurMODEL{} demonstrates good performance on task-aware prompt, }

\eat{
\eat{For example, when consider Exact Match as the evaluation metric, \OurMODEL{} achieves an average improvement of 4.02, 9.54, and 8.71 in 1shot, 2shot and 4shot scenarios compared with LLMLingua.} 
We attribute such drastic performance improvement to the following factors: (1) \OurMODEL{} retains the logical integrity of the COT example without destroying its internal structure. (2)
\OurMODEL{} benefits from the workflow that constructs the graph first, and then further compresses it. This workflow can help us compress different examples robustly without significant performance degradation on some examples.}

\eat{The No.1, No.2, No.3, No.4 shots are the different questions/examples with the inference paths. 
The GSM8K-AUG(1-shot) denotes that the No.2, No.3, and No.4 examples contain one instruction shot, i.e., No.1 shot.
GSM8K-AUG(2-shot) means that the No.3, and No.4 examples/questions contain two instruction shots (No.1 and No.2). 
GSM8K-AUG(4-shot) means that the question contain four instruction shots (No.1, No.2, No.3, No.4). 
}

\eat{
\paragraph{Implementation Details.}
In this paper, we use GPT-3.5-Turbo-0613 as
the target LLM, which can be access via OpenAI.To ensure stable and reproducible results, we adopt greedy decoding and set the temperature to 0 in all experiments.We use GPT-3.5-Turbo-0613 as our knowledge graph extraction model.

\paragraph{Datasets \& Evaluation Metric} 
We evaluate~\OurMODEL{} across three datasets, including QA task and reasoning and in-context learning (ICL) task. For QA task, we use NaturalQuestions(\citet{LostInMiddle},\citep{LostInMiddle}) and Longbench(\citet{Longbench}.\citep{Longbench}). As for reasoning and in-context learning (ICL), we use GSM8K(\citet{GSM8K},\citep{GSM8K}).

Diverging slightly from the original dataset configuration, we opt to directly extract documents containing actual answers or the closest ICL example to the problem for inclusion in the prompt, rather than incorporating all documents or all ICL examples. This approach stems from our belief that selecting a single document or single ICL example from multiple options does not accurately reflect the efficacy of prompt compression. Furthermore, it leads to an inequitable comparison between various prompt compression methodologies, given that not all such algorithms entail a document selection or an ICL example selection step from a plurality of options.

\textbf{(a) NaturalQuestions:} This benchmark comprises real-world queries from individuals utilizing Google, with responses derived from Wikipedia pages. Specifically, each query is associated with 20 pertinent documents in the initial prompt, one of which contains the correct answer. 

\textbf{(b) Longbench:} 

\textbf{(c) GSM8K:} 
}

\vspace{-1.7ex}
\section{Conclusion}
\label{sec:conclusion}
\vspace{-0.7ex}
In this work, we proposed~\OurMODEL{} that leverages graph 
structures to infer key information in the prompt in order 
to come up with a compressed prompt.
Experimental evaluation showed that~\OurMODEL{} outperforms the 
existing research on by a significant margin.
Moreover,~\OurMODEL{} addressed a key limitation of existing
prompt compression approaches, \emph{i.e.,} the compressed prompts 
are easy to read and understand for end-readers.

\newpage 
\vspace{-1.7ex}
\section{Limitations}
\label{sec:limitations}
\vspace{-0.7ex}
Some of the limitations of the \OurMODEL~ are as follows:
\begin{enumerate}
\itemsep0em   
    \item Currently our work is focused on compressing prompts that may be reformulated into structured elements, \emph{i.e.,} knowledge graph triplets $(s,r,o)$. We consider its generalization to text segments that may not be organized as graph triplets as a future research direction.
    \item The performance of~\OurMODEL{} relies on the quality of the knowledge graph constructed. We use OpenIE toolkit as our primary graph construction tool. The errors in the graph are propagated in the compressed prompt and may impact the end utility of the compressed prompt. 
\end{enumerate}

\section*{Ethics Statement}
This work fully complies with the \href{https://www.aclweb.org/portal/content/acl-code-ethics}{ACL Ethics Policy}. We declare that there are no ethical issues in this paper, to the best of our knowledge.


\bibliography{anthology,custom}

\begin{thebibliography}{26}
\providecommand{\natexlab}[1]{#1}

\bibitem[{Achiam et~al.(2023)Achiam, Adler, Agarwal, Ahmad, Akkaya, Aleman, Almeida, Altenschmidt, Altman, Anadkat et~al.}]{GPT4}
Josh Achiam, Steven Adler, Sandhini Agarwal, Lama Ahmad, Ilge Akkaya, Florencia~Leoni Aleman, Diogo Almeida, Janko Altenschmidt, Sam Altman, Shyamal Anadkat, et~al. 2023.
\newblock Gpt-4 technical report.
\newblock \emph{arXiv preprint arXiv:2303.08774}.

\bibitem[{Ali et~al.(2020)Ali, Sun, Li, and Wang}]{ali2020fine}
Muhammad~Asif Ali, Yifang Sun, Bing Li, and Wei Wang. 2020.
\newblock Fine-grained named entity typing over distantly supervised data based on refined representations.
\newblock In \emph{Proceedings of the AAAI Conference on Artificial Intelligence}, volume 34(05), pages 7391--7398.

\bibitem[{Cobbe et~al.(2021)Cobbe, Kosaraju, Bavarian, Chen, Jun, Kaiser, Plappert, Tworek, Hilton, Nakano, Hesse, and Schulman}]{GSM8K}
Karl Cobbe, Vineet Kosaraju, Mohammad Bavarian, Mark Chen, Heewoo Jun, Lukasz Kaiser, Matthias Plappert, Jerry Tworek, Jacob Hilton, Reiichiro Nakano, Christopher Hesse, and John Schulman. 2021.
\newblock \href {https://api.semanticscholar.org/CorpusID:239998651} {Training verifiers to solve math word problems}.
\newblock \emph{ArXiv}, abs/2110.14168.

\bibitem[{Dong et~al.(2022)Dong, Li, Dai, Zheng, Wu, Chang, Sun, Xu, and Sui}]{ICL}
Qingxiu Dong, Lei Li, Damai Dai, Ce~Zheng, Zhiyong Wu, Baobao Chang, Xu~Sun, Jingjing Xu, and Zhifang Sui. 2022.
\newblock A survey on in-context learning.
\newblock \emph{arXiv preprint arXiv:2301.00234}.

\bibitem[{Ji et~al.(2020)Ji, Pan, Cambria, Marttinen, and Yu}]{KGSurvey}
Shaoxiong Ji, Shirui Pan, E.~Cambria, Pekka Marttinen, and Philip~S. Yu. 2020.
\newblock \href {https://api.semanticscholar.org/CorpusID:211010433} {A survey on knowledge graphs: Representation, acquisition, and applications}.
\newblock \emph{IEEE Transactions on Neural Networks and Learning Systems}, 33:494--514.

\bibitem[{Jiang et~al.(2023{\natexlab{a}})Jiang, Wu, Lin, Yang, and Qiu}]{llmlingua}
Huiqiang Jiang, Qianhui Wu, Chin-Yew Lin, Yuqing Yang, and Lili Qiu. 2023{\natexlab{a}}.
\newblock \href {https://api.semanticscholar.org/CorpusID:263830701} {Llmlingua: Compressing prompts for accelerated inference of large language models}.
\newblock In \emph{Conference on Empirical Methods in Natural Language Processing}.

\bibitem[{Jiang et~al.(2023{\natexlab{b}})Jiang, Wu, Luo, Li, Lin, Yang, and Qiu}]{longllmlingua}
Huiqiang Jiang, Qianhui Wu, Xufang Luo, Dongsheng Li, Chin-Yew Lin, Yuqing Yang, and Lili Qiu. 2023{\natexlab{b}}.
\newblock \href {https://api.semanticscholar.org/CorpusID:263830692} {Longllmlingua: Accelerating and enhancing llms in long context scenarios via prompt compression}.
\newblock \emph{ArXiv}, abs/2310.06839.

\bibitem[{Jung and Kim(2023)}]{Jung2023DiscretePC}
Hoyoun Jung and Kyung-Joong Kim. 2023.
\newblock \href {https://api.semanticscholar.org/CorpusID:261030884} {Discrete prompt compression with reinforcement learning}.
\newblock \emph{ArXiv}, abs/2308.08758.

\bibitem[{Kim et~al.(2023)Kim, Kwon, Jo, and Choi}]{KGGPT}
Jiho Kim, Yeonsu Kwon, Yohan Jo, and Edward Choi. 2023.
\newblock \href {https://api.semanticscholar.org/CorpusID:264172465} {Kg-gpt: A general framework for reasoning on knowledge graphs using large language models}.
\newblock \emph{ArXiv}, abs/2310.11220.

\bibitem[{Kolluru et~al.(2020)Kolluru, Adlakha, Aggarwal, Mausam, and Chakrabarti}]{OpenIE}
Keshav Kolluru, Vaibhav Adlakha, Samarth Aggarwal, Mausam, and Soumen Chakrabarti. 2020.
\newblock \href {https://api.semanticscholar.org/CorpusID:222177066} {Constrained iterative labeling for open information extraction}.
\newblock In \emph{Conference on Empirical Methods in Natural Language Processing}.

\bibitem[{Lester et~al.(2021)Lester, Al-Rfou, and Constant}]{softPrompt}
Brian Lester, Rami Al-Rfou, and Noah Constant. 2021.
\newblock \href {https://api.semanticscholar.org/CorpusID:233296808} {The power of scale for parameter-efficient prompt tuning}.
\newblock In \emph{Conference on Empirical Methods in Natural Language Processing}.

\bibitem[{Lewis et~al.(2020)Lewis, Perez, Piktus, Petroni, Karpukhin, Goyal, Kuttler, Lewis, tau Yih, Rockt{\"a}schel, Riedel, and Kiela}]{RAG}
Patrick Lewis, Ethan Perez, Aleksandara Piktus, Fabio Petroni, Vladimir Karpukhin, Naman Goyal, Heinrich Kuttler, Mike Lewis, Wen tau Yih, Tim Rockt{\"a}schel, Sebastian Riedel, and Douwe Kiela. 2020.
\newblock \href {https://api.semanticscholar.org/CorpusID:218869575} {Retrieval-augmented generation for knowledge-intensive nlp tasks}.
\newblock \emph{ArXiv}, abs/2005.11401.

\bibitem[{Li(2023)}]{selectiveContext}
Yucheng Li. 2023.
\newblock \href {https://api.semanticscholar.org/CorpusID:258298489} {Unlocking context constraints of llms: Enhancing context efficiency of llms with self-information-based content filtering}.
\newblock \emph{ArXiv}, abs/2304.12102.

\bibitem[{Lin(2004)}]{rouge_2004}
Chin-Yew Lin. 2004.
\newblock \href {https://api.semanticscholar.org/CorpusID:964287} {Rouge: A package for automatic evaluation of summaries}.
\newblock In \emph{Annual Meeting of the Association for Computational Linguistics}.

\bibitem[{Liu et~al.(2023)Liu, Lin, Hewitt, Paranjape, Bevilacqua, Petroni, and Liang}]{LostInMiddle}
Nelson~F. Liu, Kevin Lin, John Hewitt, Ashwin Paranjape, Michele Bevilacqua, Fabio Petroni, and Percy Liang. 2023.
\newblock \href {https://api.semanticscholar.org/CorpusID:259360665} {Lost in the middle: How language models use long contexts}.
\newblock \emph{Transactions of the Association for Computational Linguistics}, 12:157--173.

\bibitem[{Luo et~al.(2023)Luo, Li, Haffari, and Pan}]{ReasoningOnGraph}
Linhao Luo, Yuan-Fang Li, Gholamreza Haffari, and Shirui Pan. 2023.
\newblock \href {https://api.semanticscholar.org/CorpusID:263605944} {Reasoning on graphs: Faithful and interpretable large language model reasoning}.
\newblock \emph{ArXiv}, abs/2310.01061.

\bibitem[{Meng et~al.(2022)Meng, Bau, Andonian, and Belinkov}]{Fluency}
Kevin Meng, David Bau, Alex Andonian, and Yonatan Belinkov. 2022.
\newblock \href {https://api.semanticscholar.org/CorpusID:255825985} {Locating and editing factual associations in gpt}.
\newblock In \emph{Neural Information Processing Systems}.

\bibitem[{Pan et~al.(2023)Pan, Luo, Wang, Chen, Wang, and Wu}]{KGLLMSurvey}
Shirui Pan, Linhao Luo, Yufei Wang, Chen Chen, Jiapu Wang, and Xindong Wu. 2023.
\newblock \href {https://api.semanticscholar.org/CorpusID:259165563} {Unifying large language models and knowledge graphs: A roadmap}.
\newblock \emph{ArXiv}, abs/2306.08302.

\bibitem[{Park et~al.(2023)Park, O'Brien, Cai, Morris, Liang, and Bernstein}]{Agent}
Joon~Sung Park, Joseph~C. O'Brien, Carrie~J. Cai, Meredith~Ringel Morris, Percy Liang, and Michael~S. Bernstein. 2023.
\newblock \href {https://api.semanticscholar.org/CorpusID:258040990} {Generative agents: Interactive simulacra of human behavior}.
\newblock \emph{Proceedings of the 36th Annual ACM Symposium on User Interface Software and Technology}.

\bibitem[{Sahoo et~al.(2024)Sahoo, Singh, Saha, Jain, Mondal, and Chadha}]{PromptEngineeringSurvey}
Pranab Sahoo, Ayush~Kumar Singh, Sriparna Saha, Vinija Jain, Samrat~Sohel Mondal, and Aman Chadha. 2024.
\newblock \href {https://api.semanticscholar.org/CorpusID:267636769} {A systematic survey of prompt engineering in large language models: Techniques and applications}.
\newblock \emph{ArXiv}, abs/2402.07927.

\bibitem[{Shannon(1951)}]{rongyu}
Claude~E. Shannon. 1951.
\newblock \href {https://api.semanticscholar.org/CorpusID:9101213} {Prediction and entropy of printed english}.
\newblock \emph{Bell System Technical Journal}, 30:50--64.

\bibitem[{Touvron et~al.(2023)Touvron, Lavril, Izacard, Martinet, Lachaux, Lacroix, Rozi{\`e}re, Goyal, Hambro, Azhar et~al.}]{llama}
Hugo Touvron, Thibaut Lavril, Gautier Izacard, Xavier Martinet, Marie-Anne Lachaux, Timoth{\'e}e Lacroix, Baptiste Rozi{\`e}re, Naman Goyal, Eric Hambro, Faisal Azhar, et~al. 2023.
\newblock Llama: Open and efficient foundation language models.
\newblock \emph{arXiv preprint arXiv:2302.13971}.

\bibitem[{Wei et~al.(2022)Wei, Wang, Schuurmans, Bosma, hsin Chi, Xia, Le, and Zhou}]{CoT}
Jason Wei, Xuezhi Wang, Dale Schuurmans, Maarten Bosma, Ed~Huai hsin Chi, F.~Xia, Quoc Le, and Denny Zhou. 2022.
\newblock \href {https://api.semanticscholar.org/CorpusID:246411621} {Chain of thought prompting elicits reasoning in large language models}.
\newblock \emph{ArXiv}, abs/2201.11903.

\bibitem[{Wingate et~al.(2022)Wingate, Shoeybi, and Sorensen}]{softpromptcompression}
David Wingate, Mohammad Shoeybi, and Taylor Sorensen. 2022.
\newblock \href {https://api.semanticscholar.org/CorpusID:252762169} {Prompt compression and contrastive conditioning for controllability and toxicity reduction in language models}.
\newblock In \emph{Conference on Empirical Methods in Natural Language Processing}.

\bibitem[{Xu et~al.(2023)Xu, Liu, Chen, Tang, Wang, Zhou, Hu, and Shrivastava}]{CompressThenPropmt}
Zhaozhuo Xu, Zirui Liu, Beidi Chen, Yuxin Tang, Jue Wang, Kaixiong Zhou, Xia Hu, and Anshumali Shrivastava. 2023.
\newblock \href {https://api.semanticscholar.org/CorpusID:258823240} {Compress, then prompt: Improving accuracy-efficiency trade-off of llm inference with transferable prompt}.
\newblock \emph{ArXiv}, abs/2305.11186.

\bibitem[{Yan et~al.(2023)Yan, Xu, Song, Wu, Li, and Zhang}]{Yan2023UnderstandingIL}
Jianhao Yan, Jin Xu, Chiyu Song, Chenming Wu, Yafu Li, and Yue Zhang. 2023.
\newblock \href {https://api.semanticscholar.org/CorpusID:263334398} {Understanding in-context learning from repetitions}.
\newblock \emph{ArXiv}, abs/2310.00297.

\end{thebibliography}
\bibliographystyle{acl_natbib}

\clearpage
\appendix
\section{Background}
\label{Appendix:Background}

\subsection{Knowledge Graph}
\label{sec:BCKGD_KG}
Knowledge Graph (KG) can be represented as $\mathcal{G}=\{(s,r,o)\subseteq\mathcal{E}\times\mathcal{R}\times\mathcal{E}\}$, 
where $\mathcal{E}$ and $\mathcal{R}$ denote the set of entities
and relations. Here entities are represented as nodes in $\mathcal{G}$,
while relations $\mathcal{R}$ form up edges between the nodes.

\subsection{Task-aware Prompts}
\label{Appendix:aware-prompts}
Task-aware prompts refer to the ones that are strongly related to the task and need to be re-compressed while changing the question. These prompts usually contain the specific information needed to solve the task, and some redundant parts can be removed. For example, the task can be the question, {\em "Who are the first people to win the Nobel Prize?"} and the prompt may contain a document that includes all the information about people who won the Nobel Prize. \eat{Prompts with external documentation are also an example of task-aware prompts.}

\subsection{Task-agnostic Prompts}
\label{Appendix:agnositic-prompts}
Opposite to task-aware prompts, task-agnostic prompts are weakly related to the task. This kind of prompt usually just provides LLMs with an example of what to do, such as how to solve the problem step by step. For example, the prompt may be "The weather is really nice today, emotion: positive." The model then follows this format and judges the emotion of the input sentence. \eat{Chain-of-thought Prompts are task-agnostic.}

\subsection{Chain-of-thought Prompt}
\label{Appendix:CoT}
This is a type of task-agnostic prompt. 
Chain-of-thought Prompt aims to improve performance on tasks requiring logic and calculation by mimicking human reasoning. That is $x^{info}$ include several demonstrations with detailed reasoning $p^{info}=\{p^{demo}_1,p^{demo}_2,\cdots\}$. We research how to compress chain-of-thought prompts on mathematical reasoning tasks.

\eat{\textbf{Mathematical Reasoning.} In mathematical reasoning, all questions can be solved by following the same reasoning examples. That is, $p^{info}$ is the same for all questions. We only need to compress the prompt once, regardless of the number of questions.}

\subsection{Prompt with external documentation}
\label{Appendix:EXTdoc}
This is a type of task-aware prompt. 
Prompt with external documentation implies that the prompt contains 
some additional information, such as external knowledge, that the model may refer to to answer the question. That external knowledge 
$x^{info}$ may include several external documents $p^{info}=\{p^{doc}_1,p^{doc}_2,\cdots\}$. 
Question Answering is a specific scenario for prompts with 
external knowledge.

\eat{\textbf{Question Answering} requires understanding the natural language question and then querying the external document to capture the most appropriate answer. Different questions require querying different external documents. In question answering, different questions refer to different external documents. That is, $p^{info}$ is distinct from questions. We need to perform $n$ compression on $n$ questions.}

\subsection{Token-level Prompt compression} 
In this section, we introduce the previous token-level compression method, e.g., LLMlingua~\citep{llmlingua}, 
and LongLLMlingua~\citep{longllmlingua}. 

{\bf LLMLingua} is a token-level prompt compression method that
performs compression based on perplexity. It includes a budget controller to calculate the compression ratio of demonstrations and the further compression ratio of each demonstration based on given parameters. LLMLingua uses the LLAMA-2~\cite{llama} to calculate the perplexity of each token. Finally, LLMLingua compresses the prompt based on the perplexity and compression ratio.

{\bf LongLLMlingua} is a token-level prompt compression method that aims at task-aware prompt compression. It changes the perplexity measure method and relates it to the specific question.

\eat{\section{Additional Approaches for Graph Construction}
\label{Appendix:graph-construct}
We also OpenIE~\citep{OpenIE} for graph construction. 
OpenIE is a tool proposed by Stanford University for extracting open-domain relation triples.
We use the following steps to construct the graph using OpenIE:
(i) We segment the original prompt into sentences.
(ii) We use OpenIE to extract relation triples for each sentence. Specifically, we extract $(\mathcal{E}, \mathcal{R}, \mathcal{E})$, 
where $\mathcal{E}$ represents entities, and $\mathcal{R}$ represents relations between entities.
(iii) We integrate the relation triples obtained in step (ii) 
into the form of a graph. We consider $\mathcal{E}$ as 
nodes in the graph and $\mathcal{R}$ as the edges in the graph.
Formally, it may be represented as:
\begin{equation}
    \mathcal{G} = {OpenIE}(Parser(P))
   \label{eq:Stage1}
\end{equation}
where $(P)$ is the prompt text, $Parser$ is the function that parse the prompt into sentences. OpenIE is the tool we mentioned above.
}

\section{Prompts}
\label{Appendix:prompts}
\subsection{Prompts for Graph Construction}
\label{Appendix:prompt_Graph}
\begin{tcolorbox}[colback=gray!5!white,colframe=black!75!black,title=Prompts for Graph Construction:]
\textbf{Example:}\\ 
    \textbf{Input:}\\ 
    Deadpool 2 is scheduled to be released in the United States on May 18, 2018.  A sequel, Deadpool 3, is in development.\\
    \textbf{Output:}\\
    $<$Deadpool 2; is scheduled to be released in; the United States on May 18, 2018$>$\\
$<$Deadpool 3; is in; development$>$\\
\textbf{Hint:}
\begin{itemize}
    \item You should only respond the knowledge graph triplet and not contain other word.
    \item The knowledge graph triplet is formulated as $<s, r, o>$, $s$ and $o$ should not be too long.
    \item Please keep all the relations atomic and indivisible.
\end{itemize}
Please generate the entity and relation triplets of the Input:\\
Input:{}
\end{tcolorbox}

\subsection{Instructions used for GPT-4 response Generation}
\label{Appendix:prompt_GPT4}
The instructions we used in the GPT-4 Generation are shown below:
\begin{tcolorbox}[colback=gray!5!white,colframe=black!75!black,title=Instructions used for GPT-4 response Generation:]
\textbf{Instruction1. } Condense the given paragraph to just 50\% of its original size, focusing on the core message.\\
\textbf{Instruction2. } Reduce the length of the specified paragraph to 50\%, keeping only the most essential information.\\
\textbf{Instruction3. } Compress the paragraph to 50\% of its length, ensuring the main idea is intact. Let’s do it step by step.\\
\textbf{Instruction4. } You are a prompt compression expert. Please compress the following prompt to 50\% of its original length. Let’s do it step by step.\\
\textbf{Instruction5. } You are a prompt compression expert. Please compress the following prompt with the following steps: (1) Find the key information of the document (2) Compress the prompt to 50\% of its original length without damaging key information. Let’s do it step by step.
\end{tcolorbox}


\eat{
\begin{algorithm}[H]
    \caption{\scshape Task-agnostic Compress}
    \label{alg:alg3}
    \begin{algorithmic}[1]
    \Require
        \item[] { \#$\delta:$ {The compress threshold}}
        \item[] { \#$\mathcal{G}:$ Graph structure of prompt}
        \item[] { \#$E:$ encoder}
        \item[] { \#$sim():$ function used to calculate similarity}
    \Ensure subgraph $\mathcal{G}_{subset}$
        \State $\mathcal{G}^\prime=\{\}$
        \For{$g_i \in \mathcal{G}$}
            \State $Sim_{max}=0$
            \For{$g \in \mathcal{G}^\prime$}
                \State $sim_{max}=max(sim_{max},sim(E(g),E(g_i)))$
            \EndFor
            \If{$sim_{max}<=\delta$}
            \State $\mathcal{G^\prime}.insert(g_i)$
            \EndIf
        \EndFor
        \State \Return $\mathcal{G}_{subset}$
    \end{algorithmic}
\end{algorithm}
}
\section{Experimental Details}
\label{Appendix:EXP}

\subsection{Dataset}
\label{Appendix:data}
We provide a detailed description of the evaluation data sets below.
The statistics of the dataset are given in Table~\ref{tab:DataStat}.


\begin{table*}[h]
\centering
\scalebox{0.85}{
\begin{tabular}{ccccccc}
\toprule
\multirow{2}{*}{Statistics} & \multicolumn{4}{c}{\OurDATA{}}     & \multirow{2}{*}{NaturalQuestions} & \multirow{2}{*}{ShareGPT} \\ \cline{2-5}
                            & 1shot & 2shot & 4shot & 8shot &                                   &                           \\ \hline
Token number of prompt      & 306   & 612   & 1224  & 2365  & 3040                              & 562                       \\ \hline
Number of questions         & \multicolumn{4}{c}{1319}      & 2654                              & 575                       \\ 
\bottomrule
\end{tabular}
}
\caption{The statistics of the dataset}
\label{tab:DataStat}
\end{table*}

\noindent{\bf (i) \OurDATA{}.} 
we use~\OurDATA{}, an extended version 
of original GSM8k data set allowing computations for under 
i-shot settings, \emph{i.e.,} i = \{1, 2, 4 and 8\}-shots.
Its process-flow is explained in Section~\ref{sec:data_aug}.
The statistics of the data set is shown in Table~\ref{tab:DataStat}.

\noindent{\bf (ii) NaturalQuestions.} 
It is a QA dataset that is comprised of real-world queries collected by individuals~\cite{LostInMiddle}. 
Each question of this dataset has 20 related documents, one of which 
contains the correct answer. We select documents containing answers 
as compression targets to examine better the compression performance of 
different methods on a single document.
\eat{10th, 15th, and 20th
Each example comprises query google.com query and responses derived from Wikipedia pages.
\fixchen{Cheng, plz fix the information in this para.}}

\noindent{\bf (iii) ShareGPT.} 
It is a conversation dataset encompassing users' 
conversation with ChatGPT\footnote{\url{ShareGPT.com}}.
Each data sample is a complete conversation between 
the user and ChatGPT, covering multiple languages across 
different scenarios.
Following~\citet{selectiveContext} and~\citet{llmlingua}, 
we use a subset of 575 samples provide 
by~\citet{selectiveContext} for evaluation.
We use all dialogues except the final round as the 
prompt, and the human's question in the last round 
as the question.

\subsection{Baseline Models}
\label{Appendix:Baseline}
{\bf (i) Selective-Context.}
Selective-Context by~\citet{selectiveContext} uses a small language model 
to calculate the self-information in the prompt and then filter out on 
token-level based on the self-information of each token.

\noindent{\bf (ii) LLMLingua.}
LLMLingua by~\citet{llmlingua} perform token-level prompt compression 
based on the perplexity calculated by the small language model.

\noindent{\bf (iii) LongLLMLingua.} Based on LLMLingua, LongLLMlingua by~\citet{longllmlingua} further adds a coarse-grained filtering module, which is more suitable for long document compression.

We followed their original experimental setting, uses  LLMLlingua on GSM8K and  
LongLLMlingua on NaturalQuestions. 
 
\noindent\textbf{(iv) GPT-4.} We designed five sets of prompts for 
GPT-4~\citep{GPT4} to inspire its ability on prompt compression 
and reported the best scores. Appendix~\ref{Appendix:prompts} 
displays the prompts we employed.

\subsection{Evaluation Metrics} 
\label{Appendix:Eval} 
Detailed description and mathematical formulation of the evaluation metrics  
is provided as follows: 

\noindent{\bf Exact Match (EM).} 
In EM, when the model output answer is completely consistent with the golden answer, the answer is considered correct. It is shown below.
\begin{equation}
\mathbbm{1}\left[\bigvee_{q \in \mathcal{Q}} [f(compress(q)) = q^*]\right]
\end{equation}
Where $f(\cdot)$ represents the model used to answer the question, $\mathcal{Q}$ and $q^*$ represent the question, and $q^*$ indicate the answer for question, and $compress$ indicate the prompt compression method.\\

\noindent{\bf Span Accuracy (SAcc).} We follow previous work and use SAcc to measure the performance of QA datasets. SAcc determines whether the standard answer is part of the response answer of the GPT model, as shown below.
\begin{equation}
\mathbbm{1}\left[\bigvee_{q \in \mathcal{Q}} [ q^* \in f(compress(q))]\right]
\end{equation}

\noindent{\bf Rouge.}
To measure the similarity between the output of the 
original prompt and the compressed prompt, we apply 
commonly used overlap metric ROUGE~\citep{rouge_2004}.
We report uni-gram and bi-gram overlap as the metric 
of assessing informativeness (Rouge-1 and Rouge-2), 
and the longest common sub-sequence as the metric 
of assessing ﬂuency (Rouge-L). 


\noindent{\bf Fluency (FL).}
We use fluency as a metric to measure the readability and grammatical coherence 
of the compressed prompt. Following~\citet{Fluency}, we use the following formula to 
compute the fluency.

\begin{equation}
    \text{FL}=-\sum_{k}f(k)\log_{2}f(k)
\end{equation}
where ${f(\cdot)}$ means the $n$-gram frequency distribution.


\section{Additional Results}
\label{Appendix:Analyses}



\subsection{Interpretability of Compressed prompts (Examples)}
\label{Appendix:exp-readability}
In this section, we report some examples prompts 
along with prompts compressed by~\OurMODEL{} and LLMLingua~\citep{llmlingua} for a quick 
comparison in terms of readability and/or interpretability of the compressed prompt.
As an example, for the compressed prompt text compressed using LLMLingua in Table~\ref{Appendix:ex1}, 
the text {"\em won twoes forg01 theate women prize:ertMayer" } is hard to interpret for humans.
On the contrary, the prompt compressed by~\OurMODEL{} yields comprehensive
information units helpful that are not only easy to interpret but are also \eat{. And we can easily know that {"\em Wilhelm Conrad R  ̈ontgen awarded first Nobel Prize in Physics 1901"} is }highly relevant to the question.

\onecolumn

\begin{tcolorbox}[colback=gray!5!white,colframe=black!75!black,title=Example 1.]
\label{Appendix:ex1}
\textbf{Original Prompt:}\\
Write a high-quality answer for the given question using only the provided search results.\\Document [1](Title: List of Nobel laureates in Physics) The first Nobel Prize in Physics was awarded in 1901 to Wilhelm Conrad Röntgen, of Germany, who received 150,782 SEK, which is equal to 7,731,004 SEK in December 2007.  John Bardeen is the only laureate to win the prize twice—in 1956 and 1972. Maria Skłodowska-Curie also won two Nobel Prizes, for physics in 1903 and chemistry in 1911. William Lawrence Bragg was, until October 2014, the youngest ever Nobel laureate; he won the prize in 1915 at the age of 25. Two women have won the prize: Curie and Maria Goeppert-Mayer (1963). As of 2017, the prize has been awarded\\Question: who got the first nobel prize in physics.\\Answer:\\

\textbf{Compressed Prompt by LLMLingua:}\\
Write a high-quality answer for the given question using only the provided search results.\\1Title: List of Nobelates in The first Prize1 Wilhelmrad, of who received82 in en prize. won twoes forg01 theate women prize:ertMayer (1963). As of 2017, the prize has been awarded\\Question: who got the first nobel prize in physics.\\Answer:\\

\textbf{Compressed Prompt by \OurMODEL{}:}\\
Write a high-quality answer for the given question using only the provided search results.\\Wilhelm Conrad Röntgen awarded first Nobel Prize in Physics 1901.William Lawrence Bragg won Nobel Prize in Physics 1915.Maria Goeppert-Mayer won Nobel Prize in Physics 1963\\Question: who got the first nobel prize in physics.\\Answer:
\end{tcolorbox}
\clearpage

\begin{tcolorbox}[colback=gray!5!white,colframe=black!75!black,title=Example 2.]
\label{Appendix:ex2}
\textbf{Original Prompt:}\\
Write a high-quality answer for the given question using only the provided search results.\\Document [1](Title: Distilled beverage) The term \"spirit\" in reference to alcohol stems from Middle Eastern alchemy.  These alchemists were more concerned with medical elixirs than with transmuting lead into gold. The vapor given off and collected during an alchemical process (as with distillation of alcohol) was called a spirit of the original material.\\Question: where did the term spirits for alcohol come from\\Answer:\\

\textbf{Compressed Prompt by LLMLingua:}\\
Write a high-quality answer for the given question using only the provided search results.\\) was called a spirit of the original material.\\Question: where did the term spirits for alcohol come from\\Answer:\\

\textbf{Compressed Prompt by \OurMODEL{}:}\\
Write a high-quality answer for the given question using only the provided search results .\\Alchemical process involves distillation of alcohol.Spirit stems from Middle Eastern alchemy\\Question: where did the term spirits for alcohol come from\\Answer:

\end{tcolorbox}

\begin{tcolorbox}
[colback=gray!5!white,colframe=black!75!black,title=Example 3.]
\label{Appendix:ex3}
\textbf{Original Prompt:}\\
Write a high-quality answer for the given question using only the provided search results.\\
Document [1](Title: OPEC) Organization of the Petroleum Exporting Countries 
(OPEC, OH-pek, or OPEP in several other languages) is an intergovernmental organization of 14 nations as of February 2018, founded in 1960 in Baghdad by the first five members (Iran, Iraq, Kuwait, Saudi Arabia, and Venezuela), and headquartered since 1965 in Vienna, Austria. As of 2016, the 14 countries accounted for an estimated 44 percent of global oil production and 73 percent of the world's \"proven\" oil reserves, giving OPEC a major influence on global oil prices that were previously determined by American-dominated multinational oil companies.\\
Question: how many countries are a part of opec\\
Answer:\\

\textbf{Compressed Prompt by LLMLingua:}\\
Write a high-quality answer for the given question using only the provided search results.\\
Title: OPE ofC, /\u02c8kkPEP in otheral1 nations as1 in by Venezuela. of the4  on by Americanatedinational oil companies.\\
Question: how many countries are a part of opec\\
Answer:\\

\textbf{Compressed Prompt by \OurMODEL{}:}\\
Write a high-quality answer for the given question using only the provided search results.\\Organization of the Petroleum Exporting Countries abbreviation OPEC.Organization of the Petroleum Exporting Countries nations involved 14.Organization of the Petroleum Exporting Countries founded in 1960\\Question: how many countries are a part of opec\\Answer:\\
\end{tcolorbox}

\begin{tcolorbox}
[colback=gray!5!white,colframe=black!75!black,title=Example 4.]
\label{Appendix:ex4}
\textbf{Original Prompt:}\\
Write a high-quality answer for the given question using only the provided search results.\\Document [1](Title: Subcutaneous injection) A subcutaneous injection is administered as a bolus into the subcutis, the layer of skin directly below the dermis and epidermis, collectively referred to as the cutis.  Subcutaneous injections are highly effective in administering vaccines and medications such as insulin, morphine, diacetylmorphine and goserelin. Subcutaneous, as opposed to intravenous, injection of recreational drugs is referred to as \"skin popping\". Subcutaneous administration may be abbreviated as SC, SQ, sub-cu, sub-Q, SubQ, or subcut. Subcut is the preferred abbreviation for patient safety.\\Question: where would a subcutaneous injection be made in the skin\\Answer:\\

\textbf{Compressed Prompt by LLMLingua:}\\
Write a high-quality answer for the given question using only the provided search results.\\Document [1](Title: Subcutaneous injection) A subcutaneous injection is administered as a bolus into the subcutis, the layer of skin directly below the dermis and epidermis, collectively referred to as the cutis.  Subcutaneous injections are highly effective in administering vaccines and medications such as insulin, morphine, diacetylmorphine and goserelin. Subcutaneous, as opposed to intravenous, injection of recreational drugs is referred to as \"skin popping\". Subcut SubQ, or subcut. Subcut is the preferred abbreviation for patient safety\\Question: where would a subcutaneous injection be made in the skin\\Answer:\\

\textbf{Compressed Prompt by \OurMODEL{}:}\\
Write a high-quality answer for the given question using only the provided search results.\\Subcutaneous injection administered as bolus into the subcutis.Subcutaneous injection administered for vaccines and medications.Subcutaneous injection referred to as \"skin popping\"\\Question: where would a subcutaneous injection be made in the skin\\Answer:\\
\end{tcolorbox}

\eat{
\eat{\subsection{\OurMODEL{} Results via Graph construction in~\ref{Appendix:graph-construct}}
\label{Appendix:exp-varyGraph}
\fixchen{Add some results for different graph construction approaches here..!}} 
\begin{tcolorbox}
[colback=gray!5!white,colframe=black!75!black,title=Example 1.]
Write a high-quality answer for the given question using only the provided search results.\\Gallbladder is a small hollow organ.Bile is stored in Gallbladder.Gallbladder lies beneath liver.Gallbladder is in vertebrates.Bile is concentrated in Gallbladder.Gallbladder releases bile via common bile duct.\\Question: where is gall bladder situated in human body\\Answer:
\end{tcolorbox}
\begin{tcolorbox}
[colback=gray!5!white,colframe=black!75!black,title=Example 2.]
Write a high-quality answer for the given question using only the provided search results.\\Patriots met Eagles in Super Bowl LII.Eagles won against Patriots\\Question: when's the last time the philadelphia eagles played the new england patriots\\Answer:
\end{tcolorbox}}



\end{document}